\documentclass{article}

\usepackage{microtype}
\usepackage{graphicx}
\usepackage{subfig}
\usepackage{booktabs} 
\usepackage{multirow}

\makeatletter
\def\eg{\emph{e.g.}}
\def\ie{\emph{i.e.}}
\def\aka{\emph{a.k.a.}}

\usepackage{colortbl}
\definecolor{Gray}{gray}{0.9}
\definecolor{myblue}{rgb}{0.19, 0.60, 0.60} 
\definecolor{myred}{rgb}{0.95, 0.15, 0.15} 

\usepackage{hyperref}

\usepackage[accepted]{icml2024}

\usepackage{amsmath}
\usepackage{amssymb}
\usepackage{mathtools}
\usepackage{amsthm}

\usepackage[capitalize,noabbrev]{cleveref}

\theoremstyle{plain}

\theoremstyle{definition}

\theoremstyle{remark}

\usepackage[textsize=tiny]{todonotes}

\icmltitlerunning{Realistic Unsupervised CLIP Fine-tuning with Universal Entropy Optimization}

\begin{document}

\twocolumn[
\icmltitle{Realistic Unsupervised CLIP Fine-tuning with Universal Entropy Optimization}

\icmlsetsymbol{equal}{*}

\begin{icmlauthorlist}
\icmlauthor{Jian Liang}{nlpr,ucas}
\icmlauthor{Lijun Sheng}{nlpr,ustc}
\icmlauthor{Zhengbo Wang}{nlpr,ustc}
\icmlauthor{Ran He}{nlpr,ucas}
\icmlauthor{Tieniu Tan}{nlpr,ucas,nju}
\end{icmlauthorlist}

\icmlaffiliation{nlpr}{NLPR \& MAIS, Institute of Automation, Chinese Academy of Sciences, China}
\icmlaffiliation{ustc}{University of Science and Technology of China}
\icmlaffiliation{ucas}{School of Artificial Intelligence, University of Chinese Academy of Sciences}
\icmlaffiliation{nju}{Nanjing University}

\icmlcorrespondingauthor{Jian Liang}{liangjian92@gmail.com}

\icmlkeywords{CLIP, Unsupervised fine-tuning, OOD detection}

\vskip 0.3in
]

\printAffiliationsAndNotice{}

\begin{abstract}
The emergence of vision-language models, such as CLIP, has spurred a significant research effort towards their application for downstream supervised learning tasks.
Although some previous studies have explored the unsupervised fine-tuning of CLIP, they often rely on prior knowledge in the form of class names associated with ground truth labels.
This paper explores a realistic unsupervised fine-tuning scenario, considering the presence of out-of-distribution samples from unknown classes within the unlabeled data.
In particular, we focus on simultaneously enhancing out-of-distribution detection and the recognition of instances associated with known classes.

To tackle this problem, we present a simple, efficient, and effective approach called Universal Entropy Optimization (UEO).
UEO leverages sample-level confidence to approximately minimize the conditional entropy of confident instances and maximize the marginal entropy of less confident instances.
Apart from optimizing the textual prompt, UEO incorporates optimization of channel-wise affine transformations within the visual branch of CLIP.
Extensive experiments across 15 domains and 4 different types of prior knowledge validate the effectiveness of UEO compared to baseline methods.
The code is publicly available at \url{https://github.com/tim-learn/UEO}.
\end{abstract}

\section{Introduction}
Vision-language models (VLMs) \citep{radford2021learning,li2022blip,jia2021scaling,li2022supervision} pre-trained on web-scale image-text pairs have exhibited robust zero-shot prediction capabilities, which have recently attracted increasing attention from the research community.
As an example, Contrastive Language-Image Pretraining (CLIP) \citep{radford2021learning} leverages a contrastive objective to obtain a modality-agnostic embedding space in which the paired images and texts are pulled closer and unpaired images and texts are pushed apart.
Then zero-shot classification is performed by matching the embeddings of test images and prompt-based textual descriptions (\eg, ``a photo of a [CLASS]'' and ``a picture of a [CLASS]''), merely requiring the names of all the semantic classes in downstream tasks.

Apart from the extensive research dedicated to the pre-training stage, numerous studies \citep{zhou2022learning,zhang2022tip,bahng2022exploring} have concentrated on adapting VLMs to specific downstream tasks by using task-specific labeled data.
This fine-tuning paradigm empowers VLMs to bridge both data and task gaps, leading to improved performance in recognition tasks.
In addition to multi-class classification, these pioneering strategies have also been harnessed in a spectrum of computer vision tasks, including ordinal regression \citep{li2022ordinalclip}, point cloud understanding \citep{zhang2022pointclip}, and dense prediction \citep{rao2022denseclip}.
When considering fine-tuning setups, most efforts have primarily revolved around fully supervised and few-shot supervised learning scenarios.
To pursue annotation efficiency and scalability, several recent studies \citep{huang2022unsupervised,shu2022test,tanwisuth2023pouf} have delved into the realm of unsupervised fine-tuning, remarkably achieving performance on par with few-shot supervised approaches. 
However, they still demand a priori knowledge of class names associated with ground truth labels, limiting their applicability in diverse real-world scenarios.

\begin{figure*}[h]
  \centering
  \includegraphics[width=0.95\linewidth]{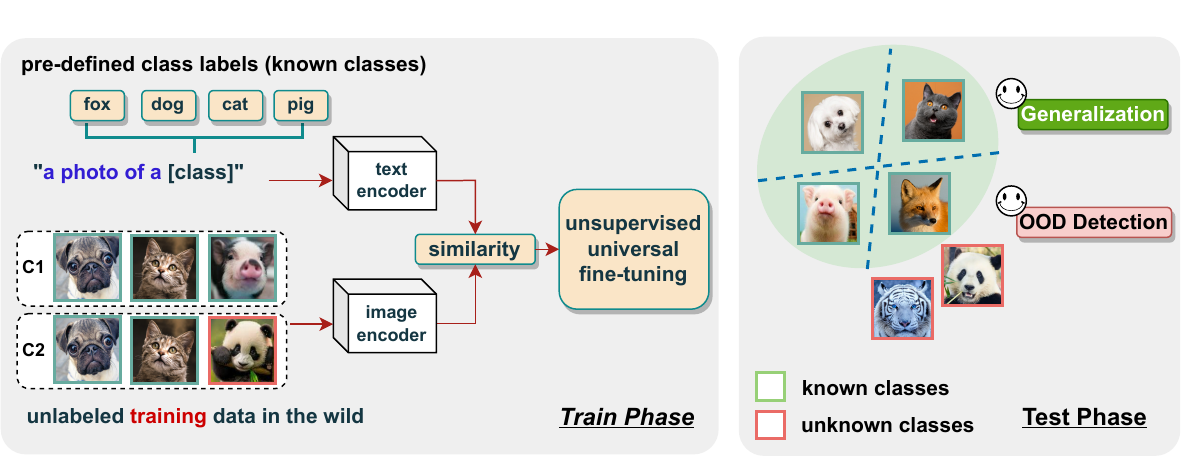}
  \caption{The basic setup of Unsupervised Universal Fine-Tuning (U$^2$-FT). During the training phase, U$^2$-FT fine-tunes the pre-trained CLIP with unlabeled in-the-wild training data according to an imprecise predefined list of class names (where `fox' may be absent in \textbf{C1} and `panda' may be included in \textbf{C2}). Note that, \textbf{a training-independent data set} containing both known classes and unknown classes (OOD) in the test phase is employed to evaluate performance in both generalization and OOD detection.}
  \label{fig:task}
\end{figure*}

To circumvent the limitation, this paper explores a novel fine-tuning setup, termed Unsupervised Universal Fine-Tuning (U$^2$-FT) with CLIP, wherein the predefined list of class names may partially overlap with the ground truth label space of unlabeled training data.
To illustrate, as depicted in Fig.~\ref{fig:task}, consider an unlabeled data set \textbf{[C2]} comprising samples from three classes (\ie, `dog', `cat', and `panda').
However, the provided predefined list of class names might be imprecise, containing four classes (\ie, `fox', `dog', `cat', `pig').
Generally, U$^2$-FT requires the fine-tuned model to demonstrate superior performance to the original CLIP in two aspects, namely, recognizing samples from classes within the predefined list (`dog' and `cat'), as well as identifying samples from classes not present in that list (`panda'), commonly referred to as out-of-distribution (OOD) samples.
Given the potential scarcity of OOD samples \textbf{[C1]}, U$^2$-FT evaluates both generalization and OOD detection through a new test data set encompassing both OOD samples and samples from classes within the predefined list.
Typically, U$^2$-FT poses two primary technical challenges for designing fine-tuning strategies: (1) fitting the entire data with CLIP will deteriorate the ability to detect OOD samples due to the potential presence of OOD samples, and (2) matching the label distribution of unlabeled data with the pre-defined one can be risky due to the potential absence of certain classes.

We propose to address the challenges by presenting a parameter-efficient approach termed Universal Entropy Optimization (UEO).
UEO aims to minimize non-OOD samples' information entropy while maximizing the OOD samples' entropy.
Since we do not know which samples are OOD, UEO readily utilizes the confidence of unlabeled data in CLIP as sample-level weight.
To avoid the potential risks associated with OOD sample exposure through entropy maximization, UEO employs a reverse weighting strategy to aggregate the predictions first, before subsequently maximizing the marginal entropy.
Besides, UEO takes into account the optimization of channel-wise affine transformations in the image encoder of CLIP, in addition to the textual prompt, to ensure parameter efficiency.
Overall, UEO is remarkably simple, requiring alterations to only a few lines of code.
Our contributions are summarized as follows: 
(1). We introduce a new unsupervised fine-tuning setup with CLIP that requires minimal prior knowledge of the label space for unlabeled data.
(2). The proposed UEO devises a new parameter-efficient strategy and elegantly incorporates the sample-level confidence during entropy optimization. 
(3). Extensive experiments demonstrate that UEO surpasses existing unsupervised methods in improving both generalization and OOD detection performance.

\section{Related Work}
\subsection{Prompt Tuning of CLIP}
Recent efforts have been devoted to exploring transfer learning with VLMs \citep{chen2023vlp,wang2023large} to enhance their capacity to generalize to specific downstream tasks \citep{zhang2024vision}.
Prompt tuning \citep{lester2021power}, initially devised for adapting language models, optimizes the text embedding space while leaving the foundational model parameters unchanged.
This technique has gained significant interest in fine-tuning VLMs (\eg, CLIP) for vision tasks \citep{zhou2022conditional,zhou2022learning,zhu2023prompt} owing to its parameter-efficient nature.
For example, CoOp \citep{zhou2022learning} employs prompt tuning within the language branch of CLIP by utilizing a limited amount of labeled data, while VPT \citep{bahng2022exploring} introduces visual prompts by modifying the pixels of images.
To offer greater versatility, both UPT \citep{zang2022unified} and MaPLE \citep{khattak2023maple} propose to optimize prompts within both the vision and language branches.

Depending on the availability of annotations during the learning process, current fine-tuning setups fall within three main categories: fully supervised transfer \citep{rao2022denseclip,bahng2022exploring,wortsman2022robust}, few-shot supervised transfer \citep{zhou2022learning,zhou2022conditional,zhang2022tip,wang2023improving}, and unsupervised transfer \citep{huang2022unsupervised,shu2022test,li2023masked,tanwisuth2023pouf}.
Unlike the other two transfer setups that depend on labeled downstream data, unsupervised transfer only harnesses unlabelled downstream data for fine-tuning CLIP.
This paradigm sounds more challenging but also more promising and efficient for adapting CLIP, as it does not require costly annotation efforts and can better capture the underlying data distribution.
Note that, this paper extends the scope of existing unsupervised transfer methods by assuming that not all unlabeled data belongs to classes within a predefined list.

\subsection{OOD Detection of CLIP}
Out-of-distribution (OOD) detection \citep{yang2021generalized} focuses on identifying instances or examples unrelated to the in-distribution task, which is critical for the real-world deployment of machine learning models.
However, exploring CLIP for OOD detection remains an interesting but relatively recent research topic, with only a handful of previous efforts within this field \citep{fort2021exploring,esmaeilpour2022zero,ming2022delving,wang2023clipn}.
Unlike \citep{fort2021exploring,esmaeilpour2022zero}, which rely on prior information on OOD samples, MCM \citep{ming2022delving} offers a training-free OOD detection approach that measures the similarities between visual features with textual concepts.
Besides, a recent study \citep{liao2023r} fine-tunes CLIP using labeled data to improve both generalization and OOD detection performance, even incorporating novel words from WordNet.
Two concurrent studies \citep{ming2024does,miyai2023locoop} investigate the performance of fine-tuned CLIP after few-shot in-distribution (ID) classification.
In contrast, UEO enhances the OOD detection performance of CLIP using only unlabeled data, without any additional information apart from a predefined list of class names.

\subsection{Unsupervised Model Adaptation}
Unsupervised model adaptation (\aka, source-free domain adaptation) \citep{liang2020we,li2020model,huang2021model,kundu2022balancing} has emerged as a popular research topic in transfer learning, intending to transfer a well-trained model from a labeled source domain to an unlabeled but related target domain.
As classified in a recent survey paper \citep{liang2023comprehensive}, unsupervised model adaptation methods can be broadly categorized into four popular schemes: pseudo-labeling, consistency training, clustering-based training, and source distribution estimation.
However, these methods typically require a closely related source domain to train the source model, limiting their practical applicability in real-world scenarios.
Conversely, by employing CLIP, we effortlessly acquire a high-quality source model with the help of class names from the target task. 

While most research in unsupervised model adaptation has focused on closed-set scenarios, a few studies \citep{liang2021umad,feng2021open,qu2023upcycling} have also investigated open-set model adaptation, which deals with target tasks that contain additional classes not present in the source task.
In such cases, these novel classes are treated as a distinct `unknown' category, and their accuracy is evaluated accordingly.
Nonetheless, existing model adaptation techniques are always tailored to address a single category shift scenario, specifically, open-set transfer \citep{kundu2020towards,feng2021open,liang2021umad,qu2023upcycling} or closed-set transfer \citep{liang2020we,li2020model,huang2021model,liang2022source}.
To the best of our knowledge, DANCE \citep{saito2020universal} is currently the only domain adaptation method that can handle different types of category shift scenarios (\ie, closed-set  \citep{zhang2021unsupervised}, partial-set \citep{cao2018partial}, open-set, and open-partial-set \citep{you2019universal}) without prior knowledge of the specific scenario.
Building upon the universal domain adaptation concept of DANCE, this paper presents the problem of unsupervised universal fine-tuning, which adapts CLIP to unlabeled training data with the presence of an imprecise predefined list of class names.
Notably, we employ OOD detection to distinguish between known classes and OOD classes, thereby avoiding the need for sensitivity thresholds.

\section{Unsupervised Universal Fine-Tuning}
\subsection{Preliminary}
In this paper, we employ CLIP \citep{radford2021learning} as a representative VLM for unsupervised universal fine-tuning throughout this paper since it is a pioneering work that has led to significant advancements in various computer vision tasks.
For the sake of simplicity, we focus on the image classification task.
Typically, the CLIP model adopts a straightforward dual-stream architecture with an image encoder, denoted as $g_I(\cdot)$, and a text encoder, denoted as $g_T(\cdot)$.
Each encoder processes input data from the corresponding modality.
In its pre-training phase, CLIP leverages a self-supervised contrastive objective to learn image-text correspondences from noisy image-text pairs sourced from the Internet.
As a result, the features of paired images and texts are close to each other in the shared embedding space.

\begin{figure*}[!htb]
    \centering
    \small
    \setlength\tabcolsep{1mm}
    \renewcommand\arraystretch{0.1}
    \begin{tabular}{ccc}
        \includegraphics[width=0.3\linewidth]{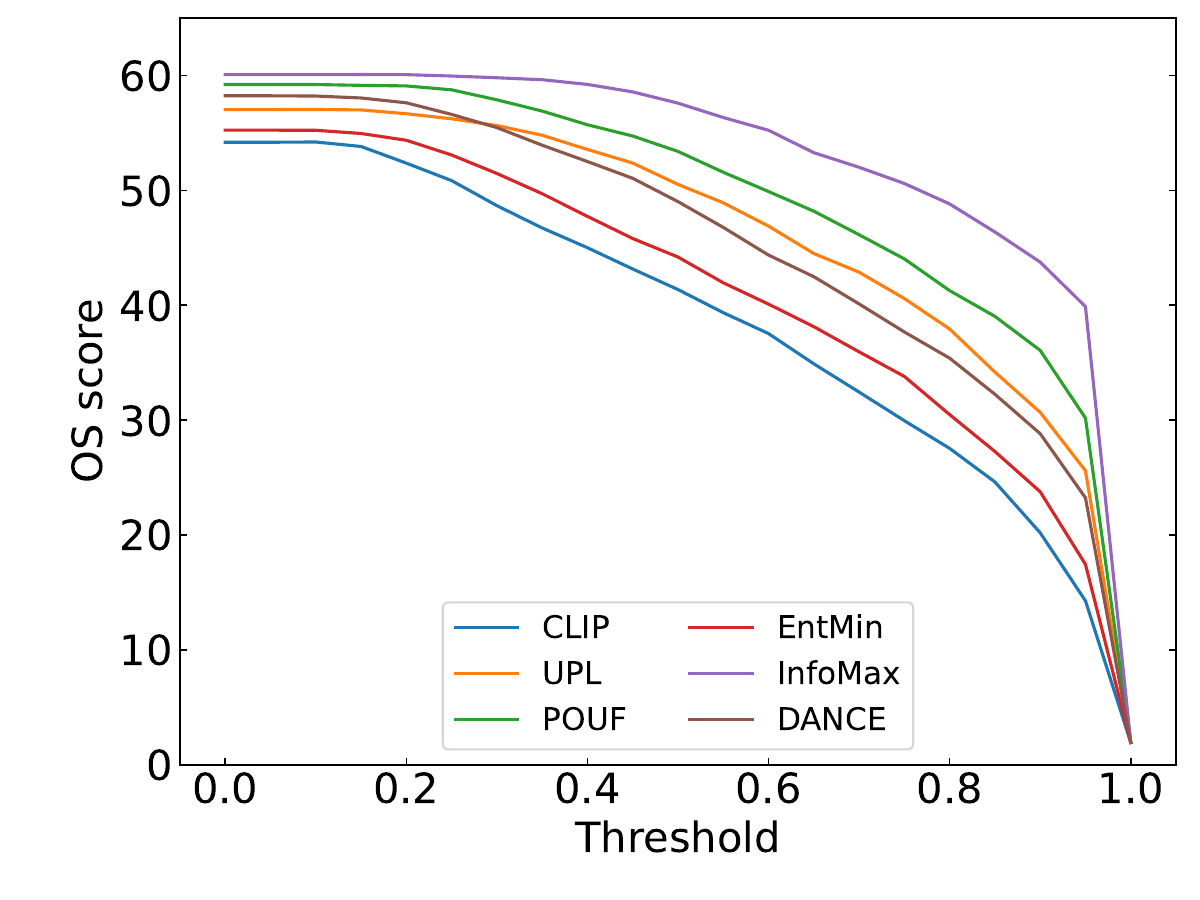} &
        \includegraphics[width=0.3\linewidth, clip]{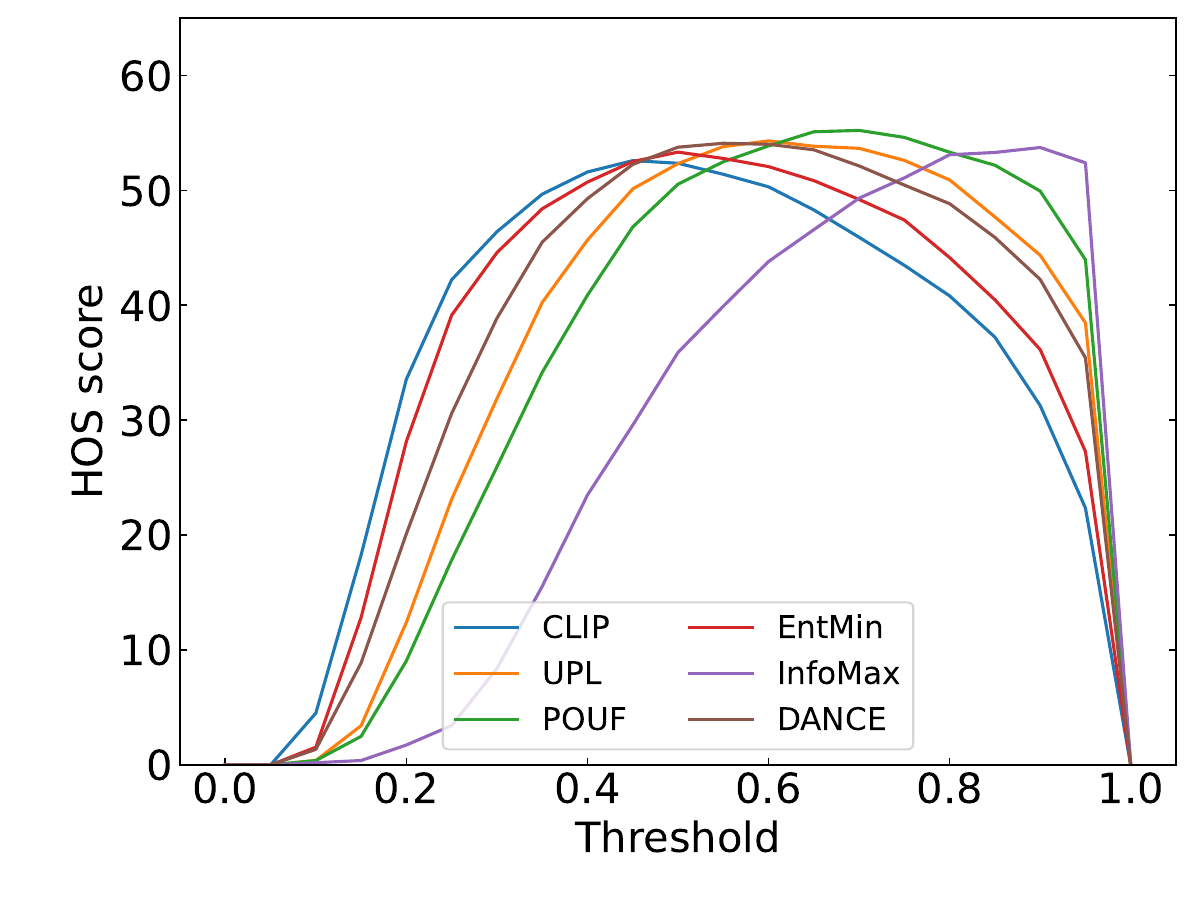} &
        \includegraphics[width=0.3\linewidth]{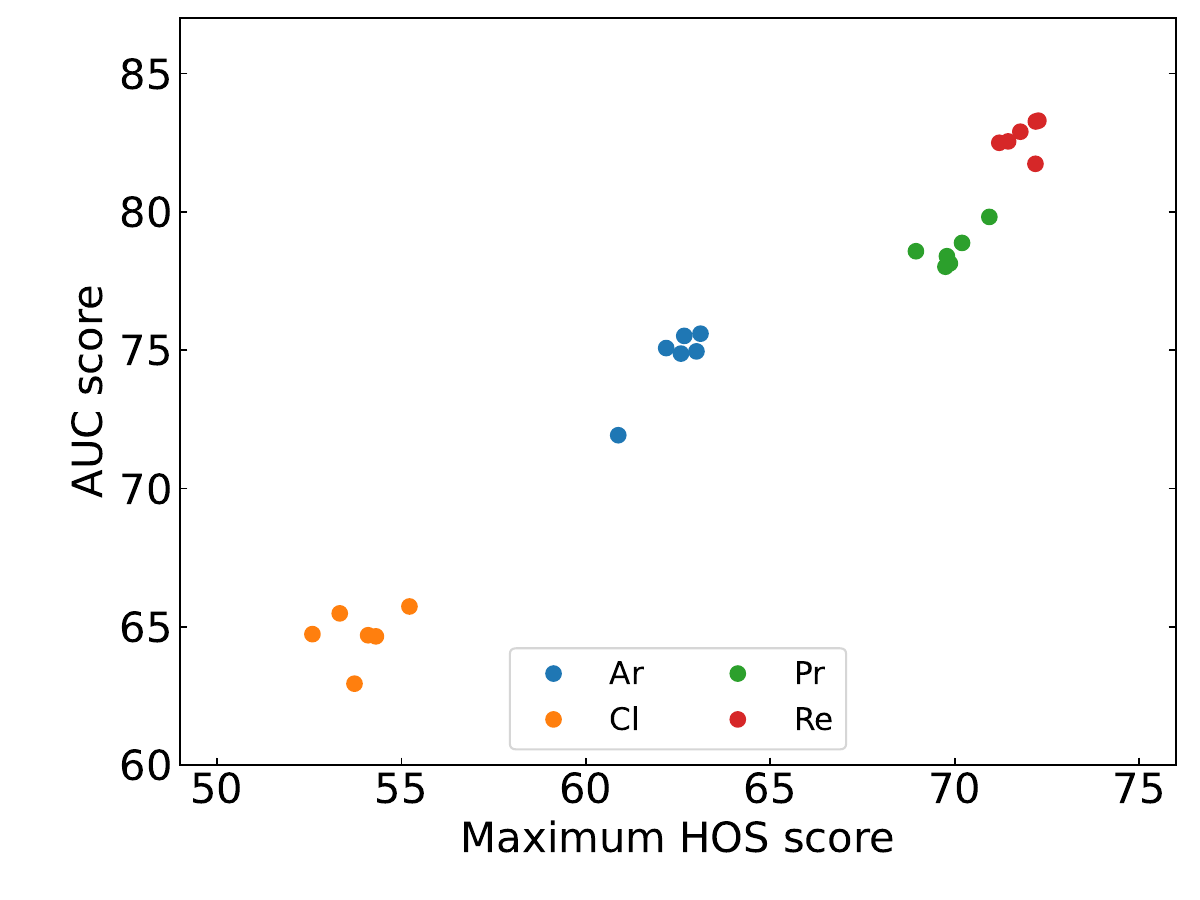} \\
        (a) OS under different thresholds & (b) HOS under different thresholds & (c) AUC v.s. the maximum HOS score
    \end{tabular}
    \caption{OS and HOS scores \citep{bucci2020effectiveness} of different methods with the change of threshold under open-partial category shift on the Cl domain of OfficeHome \citep{venkateswara2017deep} are shown in (a-b). The relationship between AUC and the maximum HOS score is depicted in (c) for four different domains.}
    \label{fig:eval}
\end{figure*} 

To facilitate zero-shot prediction in downstream tasks, CLIP generates a prompt (\eg, ``a photo of a [CLASS]'') for each class by replacing the [CLASS] token with the name of the corresponding class.
This technique aims to reduce the gap between the text distribution of the pre-training dataset and that of the target downstream task.
Then, we get the text embedding of each class encoded by the text encoder ${\{\mathbf{T}_c\}}_{c=1}^C$, where $C$ denotes the number of classes in the target task.
For making predictions, we compare the image embedding $\mathbf{I}_x=g_I(x)$ of an input image $x$ against a set of text embeddings ${\{\mathbf{T}_c\}}_{c=1}^C$ and obtain the probability that $x$ belongs to class $c$ using a softmax operation:
\begin{equation} 
\label{eq:clip}
p_c(x) = p(y=c|x) = \frac{\exp(\mathbb{S}(\mathbf{I}_x, \mathbf{T}_c)/ \tau)} {\sum_{i=1}^{C} \exp(\mathbb{S}(\mathbf{I}_x, \mathbf{T}_i)/\tau)},
\end{equation}
where $\mathbb{S}(\cdot, \cdot)$ denotes the cosine similarity metric between embeddings, and the temperature parameter $\tau$ is set to 0.01 by default.
Note that the accuracy of zero-shot inference is highly dependent on the quality of the candidate class names $\{y_1, \dots, y_C\}$ selected for the prediction task.

In addition to its impressive zero-shot classification capabilities, CLIP has also demonstrated remarkable performance in zero-shot OOD detection as reported in \citep{ming2022delving,wang2023clipn}.
For the sake of simplicity, we adopt the maximum softmax probability score as,
\begin{equation} 
\label{eq:msp}
S(x) = \max_{c} p_c(x).
\end{equation}
Due to the strong zero-shot classification ability, ID samples will be matched to one of the textual descriptions in the candidate list with a high score, and vice versa.
Formally, a standard OOD detection function can be expressed as:
$f_{\lambda}({x})=\begin{cases}
\text{ID} & S(x) \ge \lambda \\
\text{OOD} & S(x) <\lambda
\end{cases},$
where $\lambda$ is a chosen threshold so that a high fraction of ID data is above the threshold in real-world applications.
For samples that are categorized as ID, we easily obtain the class prediction through $\hat{y} = \arg\max_c p_c(x)$.

\subsection{Problem Setting}
\label{sec:problem}
With a predetermined name list of known classes, denoted as $\{y_1,\dots, y_C\}$, the objective of Unsupervised Universal Fine-Tuning (U$^2$-FT) is to facilitate the adaptation of CLIP to unlabeled data $\mathcal{X}_t$ in the wild.
U$^2$-FT is primarily designed to enhance the performance of CLIP in two key aspects: (1) accurately classifying samples affiliated with the `known' classes from the aforementioned list (ID generalization) and (2) effectively identifying samples beyond these designated classes (OOD detection).
To better understand the wildness of unlabeled data, we denote the label space of the predefined list as $L_p$, and the label space of the unlabeled data as $L_u$.
Prior unsupervised fine-tuning methods \citep{huang2022unsupervised,tanwisuth2023pouf} only consider the closed-set category shift scenario (\ie, $L_u=L_p$).
However, three other prevalent category shift scenarios exist: partial-set ($L_u \subset L_p$), open-set ($L_p \subset L_u$), and open-partial ($L_u \cap L_p \not =\emptyset, L_u \not\subset L_p, L_p \not\subset L_u$).
Due to the unlabeled nature of downstream data, we may not know in advance which of these scenarios will occur.
Hence, we adhere to the notion of `universal' as introduced in the pioneering work of DANCE \citep{saito2020universal}, delving into the universal adaptation of CLIP to wild unlabeled data.

\textbf{Evaluation.} 
Generally, we evaluate the recognition performance after unsupervised fine-tuning across four distinct category shift scenarios.
Unlike DANCE \citep{saito2020universal}, which assesses accuracies using training unlabeled data, we opt for an independent test set comprising both ID and OOD samples.
This choice is motivated by the fact that, in scenarios involving closed-set and partial-set category shifts, the absence of OOD samples in the test set makes it unfeasible to assess the performance of OOD detection.
Furthermore, DANCE treats all OOD samples as an extra `unknown' class and calculates the accuracy averaged over all classes (\ie, OS score) \citep{panareda2017open,saito2018open}.
Meanwhile, numerous open-set domain adaptation methods embrace the HOS score \citep{bucci2020effectiveness,fu2020learning}, replacing the simple average with the harmonic mean.

Both OS and HOS scores necessitate the accuracy of the `unknown' class, which highly relies on the selected threshold $\lambda$.
As an alternative approach, we align with the prevalent practice in the field of OOD detection \citep{yang2021generalized,yang2023auto} and incorporate another widely accepted metric: the area under the receiver operating characteristic curve (AUC).
This metric is complemented by the assessment of per-class accuracy for ID samples within the test set. 
In Fig.~\ref{fig:eval}(a-b), we first illustrate the sensitivity of OS and HOS scores of different methods as the threshold varies.
Clearly, the highest OS score is consistently achieved when no samples are classified as OOD.
The variation in HOS score differs across different methods and is notably affected by the threshold.
In addition, as depicted in Fig.~\ref{fig:eval}(c), the examination reveals a positive linear correlation between the AUC score and the maximum HOS score, indicating the superiority of the AUC score for open-set evaluation.

\subsection{Universal Entropy Optimization (UEO)}
To adapt CLIP to the unlabeled data, we consider using Shannon entropy as the optimization objective function. 
The idea is to minimize the entropy of the model's prediction for each instance, making them closer to one of these prototypes in the feature space. 
However, when the training data contains OOD samples, entropy minimization may have the unintended effect of weakening the model's ability to reject them.
Ideally, OOD samples should exhibit dissimilarity to any of the classes in the predefined list, thereby we can use entropy maximization instead to make the model produce approximately uniform predictions.
Nevertheless, this approach becomes infeasible in unsupervised fine-tuning due to the lack of knowledge about which samples are OOD.

Previous studies \citep{ming2022delving,wang2023clipn} have shown that using the maximum softmax probability score in Eq.~(\ref{eq:msp}) for CLIP can achieve impressive performance in OOD detection.
Inspired by this, we treat such scores as sample-level weights $w(x)$ to approximately achieve entropy minimization and maximization at the same time.
Formally, the unified objective of entropy optimization becomes
\begin{equation}
\mathcal{L} = \sum_{x \in \mathcal{B}_t} \widetilde{w}(x) \mathcal{H}(p(x)) - \sum_{x \in \mathcal{B}_t} \widetilde{\Phi}(w(x)) \mathcal{H}(p(x)), 
\label{eq:went}
\end{equation}
where $\mathcal{B}_t$ denotes a mini-batch of $\mathcal{X}_t$, $\mathcal{H}(p(x)) = -\sum\nolimits_{c=1}^{C} p_c(x) \log p_c(x)$ denotes the Shannon entropy of $p(x)$, and $\Phi(\cdot)$ represents a monotonically decreasing function, such as $\Phi(w) = 1/w$.
The normalized weight within a mini-batch is defined as $\widetilde{w}(x)=\frac{w(x)}{\sum_x w(x)}$, which emphasizes confident samples during entropy minimization.
In contrast, the normalized weight before entropy maximization is denoted as $\widetilde{\Phi}(w(x))=\frac{\Phi(w(x))}{\sum_x \Phi(w(x))}$, which places emphasis on potential OOD samples.

The combination of two weighted entropy terms in Eq.~(\ref{eq:went}) may be optimal when the weights within a mini-batch exhibit significant diversity.
If no OOD samples are present in the mini-batch, the second entropy term would deteriorate the adaptation process by increasing the entropy of a difficult sample belonging to one of the targeted classes.
To mitigate these potential risks, we apply entropy maximization over the average prediction of all the OOD samples and obtain the following objective,
\begin{equation}
\mathcal{L} = \sum_{x \in \mathcal{B}_t} \widetilde{w}(x) \mathcal{H}(p(x)) - \mathcal{H}(\bar{p}),\; 
\label{eq:went2}
\end{equation}
where $\bar{p} = \sum_{x \in \mathcal{B}_t} \widetilde{\Phi}(w(x)) p(x)$ is the weighted average of predictions for each sample within the mini-batch $\mathcal{B}_t$.
Intuitively, the objective in Eq.~(\ref{eq:went}) resembles the mutual information maximization loss \citep{liang2020we}, which can also be decomposed into two entropy terms.
When all the samples within a mini-batch share the same weight, namely, $\widetilde{w}(x)=\widetilde{\Phi}(w(x))=\frac{1}{\|\mathcal{B}_t\|}$, where $\|\mathcal{B}_t\|$ denotes the batch size, the objective exactly degrades to the information maximization loss.
This indicates that even if no OOD samples exist in the unlabeled data, optimizing the second term in Eq.~(\ref{eq:went2}) can still be beneficial.

\textbf{Parameter efficiency.}
During the unsupervised adaptation process, we employ a parameter-efficient fine-tuning paradigm \citep{lialin2023scaling} for foundation models, wherein only a small number of parameters instead of the entire model are modified during the fine-tuning process.
In particular, we follow CoOp \citep{zhou2022learning} by optimizing the textual prompt, namely the learnable word vectors $\{[\text{V}_i]\}_{i=1}^m$ in the textual sentence ``[V$_1$], [V$_2$], \dots, [V$_m$], [CLASS]'', where $m$ denotes the length of textual prompt.
Several prior studies \citep{bahng2022exploring,zang2022unified,khattak2023maple} have demonstrated that the integration of visual prompts enhances the few-shot labeled adaptation of CLIP. 
However, it is worth noting that these methods are exclusively suitable for transformer-based visual branches.
Instead, we draw inspiration from TENT \citep{wang2021tent} and introduce an approach to optimize the affine parameters within the normalization layers of the image branch, complementing the optimization of the textual prompt.

\section{Experiments}
\subsection{Setup}
\textbf{Datasets.}
We employ four popular domain adaptation datasets, \ie, \textbf{Office} \citep{saenko2010adapting} that comprises 3 domains of 31 object categories, \textbf{OfficeHome} \citep{venkateswara2017deep} that encompasses 65 categories across 4 domains, \textbf{VisDA-C} \citep{peng2017visda} that includes 2 distant domains of 12 classes, and \textbf{DomainNet (DN)} \citep{peng2019moment} that contains 345 classes distributed in 6 styles.

\setlength{\tabcolsep}{3.0pt}
\begin{table*}[!htb]
\centering
\caption{Results (\%) on different domains under the closed-set category shift (ResNet-50). [\textbf{\color{myred}{Best}}/ \textbf{\color{myblue}{best}} values.]}
\label{tab:cda}
    \resizebox{0.96\textwidth}{!}{
    \begin{tabular}{l|l|cc>{\columncolor{Gray}}c|cc>{\columncolor{Gray}}c|cc>{\columncolor{Gray}}c|cc>{\columncolor{Gray}}c|cc>{\columncolor{Gray}}c}
    \toprule
        \multirow{2}{*}{\rotatebox[origin=c]{270}{OPT}}& Tasks & \multicolumn{3}{c|}{Office} & \multicolumn{3}{c|}{OfficeHome} & \multicolumn{3}{c|}{VisDA-C} & \multicolumn{3}{c|}{DomainNet} & \multicolumn{3}{c}{Average} \\ 
        & Methods & ACC & AUC & HAA & ACC & AUC & HAA & ACC & AUC & HAA & ACC & AUC & HAA & ACC & AUC & HAA \\ \midrule
        $\times$ & CLIP \citep{radford2021learning} & 73.2 & 82.8 & 77.7 & 73.2 & 75.0 & 74.0 & 88.9 & 81.6 & 85.1 & 47.3 & 66.5 & 53.0 & 64.9 & 74.0 & 67.8 \\ \midrule
        \multirow{6}{*}{\rotatebox[origin=c]{270}{Textual Prompt}}
        & EntMin \citep{wang2021tent} & 73.3 & 82.8 & 77.7 & 73.2 & 75.1 & 74.0 & 89.0 & 81.9 & 85.3 & 46.4 & 65.9 & 51.3 & 64.6 & 73.9 & 67.2 \\ 
        & InfoMax \citep{liang2020we} & 78.6 & 82.3 & 80.3 & 76.0 & 75.7 & \textbf{\color{myblue}75.8} & 91.8 & 81.2 & 86.2 & 50.6 & 67.0 & 55.6 & 68.5 & 74.3 & 70.0 \\ 
        & DANCE \citep{saito2020universal} & 74.8 & 83.2 & 78.8 & 74.5 & 75.9 & 75.1 & 92.5 & 82.1 & 86.9 & 47.2 & 65.4 & 51.6 & 66.0 & 74.0 & 68.0 \\ 
        & UPL \citep{huang2022unsupervised} & 76.8 & 82.4 & 79.4 & 75.6 & 75.7 & 75.6 & 89.0 & 81.9 & 85.3 & 50.1 & 67.3 & 55.6 & 67.4 & 74.5 & 69.6 \\ 
        & POUF \citep{tanwisuth2023pouf} & 76.2 & 83.0 & 79.4 & 75.4 & 76.3 & \textbf{\color{myblue}75.8} & 92.1 & 83.9 & \textbf{\color{myblue}87.8} & 49.9 & 66.3 & 54.8 & 67.6 & 74.6 & 69.7 \\ 
        & UEO & 77.6 & 83.5 & \textbf{\color{myblue}80.4} & 76.2 & 74.9 & 75.5 & 91.1 & 84.4 & 87.6 & 50.8 & 67.3 & \textbf{\color{myblue}55.8} & 68.3 & 74.9 & \textbf{\color{myblue}70.2} \\ 
        \midrule
        \multirow{6}{*}{\rotatebox[origin=c]{270}{+ NormLayers}}
        & EntMin \citep{wang2021tent} & 73.7 & 82.9 & 78.0 & 73.8 & 75.6 & 74.6 & 90.3 & 82.7 & 86.3 & 46.7 & 66.1 & 51.5 & 65.1 & 74.2 & 67.6 \\
        & InfoMax \citep{liang2020we} & 79.2 & 82.5 & 80.8 & 77.0 & 75.8 & \textbf{\color{myred}76.3} & 92.6 & 81.4 & 86.6 & 52.0 & 66.3 & 56.8 & 69.5 & 74.1 & 70.8 \\
        & DANCE \citep{saito2020universal} & 74.6 & 83.2 & 78.6 & 75.0 & 76.1 & 75.5 & 90.2 & 84.7 & 87.3 & 48.1 & 65.7 & 52.3 & 66.2 & 74.5 & 68.4 \\
        & UPL \citep{huang2022unsupervised} & 77.0 & 83.0 & 79.8 & 76.0 & 76.5 & 76.1 & 89.0 & 81.7 & 85.2 & 51.0 & 67.5 & 56.6 & 67.9 & 74.9 & 70.3 \\
        & POUF \citep{tanwisuth2023pouf} & 76.7 & 83.3 & 79.8 & 76.1 & 76.4 & 76.2 & 92.3 & 84.1 & 88.0 & 50.5 & 65.7 & 55.4 & 68.1 & 74.5 & 70.2 \\
        & UEO & 78.1 & 84.1 & \textbf{\color{myred}80.9} & 76.8 & 75.6 & 76.1 & 92.2 & 84.6 & \textbf{\color{myred}88.2} & 51.9 & 67.2 & \textbf{\color{myred}57.0} & 69.1 & 75.1 & \textbf{\color{myred}71.1} \\
    \bottomrule
    \end{tabular}
    }
\end{table*}

\setlength{\tabcolsep}{3.0pt}
\begin{table*}[!ht]
\centering
\caption{Results (\%) on different domains under the partial-set category shift (ResNet-50).}
\label{tab:pda}
    \resizebox{0.96\textwidth}{!}{
    \begin{tabular}{l|l|cc>{\columncolor{Gray}}c|cc>{\columncolor{Gray}}c|cc>{\columncolor{Gray}}c|cc>{\columncolor{Gray}}c|cc>{\columncolor{Gray}}c}
    \toprule
        \multirow{2}{*}{\rotatebox[origin=c]{270}{OPT}} & Tasks & \multicolumn{3}{c|}{Office} & \multicolumn{3}{c|}{OfficeHome} & \multicolumn{3}{c|}{VisDA-C} & \multicolumn{3}{c|}{DomainNet} & \multicolumn{3}{c}{Average} \\ 
        & Methods & ACC & AUC & HAA & ACC & AUC & HAA & ACC & AUC & HAA & ACC & AUC & HAA & ACC & AUC & HAA \\ \midrule
        $\times$ & CLIP \citep{radford2021learning} & 73.2 & 82.8 & 77.7 & 73.2 & 75.0 & 74.0 & 88.9 & 81.6 & 85.1 & 47.3 & 66.5 & 53.0 & 64.9 & 74.0 & 67.8 \\ \midrule
        \multirow{6}{*}{\rotatebox[origin=c]{270}{Textual Prompt}}
        & EntMin \citep{wang2021tent} & 73.3 & 82.8 & 77.7 & 73.1 & 75.1 & 74.0 & 89.0 & 81.8 & 85.2 & 46.4 & 65.9 & 51.3 & 64.6 & 73.8 & 67.1 \\ 
        & InfoMax \citep{liang2020we} & 75.6 & 83.2 & \textbf{\color{myblue}79.2} & 75.3 & 75.5 & 75.3 & 87.8 & 79.8 & 83.6 & 50.6 & 67.0 & 55.6 & 67.1 & 74.2 & 69.3 \\ 
        & DANCE \citep{saito2020universal} & 74.0 & 82.8 & 78.2 & 74.4 & 76.1 & 75.1 & 86.7 & 74.1 & 79.8 & 48.4 & 65.9 & 52.5 & 65.5 & 73.1 & 67.3 \\
        & UPL \citep{huang2022unsupervised} & 74.3 & 83.1 & 78.4 & 73.7 & 74.9 & 74.2 & 89.0 & 81.7 & 85.2 & 49.7 & 67.3 & 55.4 & 66.3 & 74.4 & 69.0 \\ 
        & POUF \citep{tanwisuth2023pouf} & 75.1 & 83.0 & 78.8 & 74.8 & 76.1 & 75.4 & 88.3 & 78.3 & 83.0 & 49.9 & 66.5 & 54.8 & 66.7 & 73.9 & 68.8 \\ 
        & UEO & 75.8 & 82.9 & \textbf{\color{myblue}79.2} & 75.9 & 75.9 & \textbf{\color{myblue}75.8} & 89.0 & 82.2 & \textbf{\color{myblue}85.5} & 50.8 & 67.3 & \textbf{\color{myblue}55.8} & 67.6 & 74.7 & \textbf{\color{myblue}69.8} \\ 
        \midrule
        \multirow{6}{*}{\rotatebox[origin=c]{270}{+ NormLayers}}
        & EntMin \citep{wang2021tent} & 73.5 & 82.8 & 77.9 & 73.7 & 75.4 & 74.4 & 89.7 & 81.8 & 85.6 & 46.7 & 66.1 & 51.5 & 65.0 & 74.0 & 67.4 \\
        & InfoMax \citep{liang2020we} & 76.7 & 83.6 & \textbf{\color{myred}80.0} & 76.3 & 75.4 & 75.8 & 88.7 & 79.4 & 83.7 & 51.7 & 66.4 & 56.7 & 68.2 & 74.0 & 70.0 \\
        & DANCE \citep{saito2020universal} & 73.6 & 82.7 & 77.9 & 74.3 & 75.5 & 74.8 & 87.6 & 79.6 & 83.4 & 48.5 & 65.8 & 52.6 & 65.6 & 73.6 & 67.7 \\
        & UPL \citep{huang2022unsupervised} & 74.6 & 83.4 & 78.7 & 74.2 & 75.6 & 74.8 & 89.2 & 81.9 & 85.4 & 50.7 & 67.4 & 56.4 & 66.9 & 74.7 & 69.6 \\
        & POUF \citep{tanwisuth2023pouf} & 75.2 & 83.2 & 79.0 & 75.8 & 76.4 & 76.0 & 89.8 & 80.9 & 85.1 & 50.3 & 66.1 & 55.4 & 67.4 & 74.2 & 69.6 \\
        & UEO & 76.4 & 83.4 & 79.7 & 76.6 & 76.3 & \textbf{\color{myred}76.4} & 90.0 & 82.9 & \textbf{\color{myred}86.3} & 51.8 & 67.2 & \textbf{\color{myred}57.0} & 68.4 & 75.0 & \textbf{\color{myred}70.6} \\
    \bottomrule
    \end{tabular}
    }
\end{table*}

\textbf{Protocols \& Evaluations.}
U$^2$-FT assesses both the generalization performance of known classes and the OOD detection capability towards samples from OOD classes, as described in Section \ref{sec:problem}.
For each category shift, we determine the label space of the training unlabeled data based on the specific shift and select all the data from these classes.
The detailed data splits about known classes and unknown classes can be found in Appendix~\ref{app:split}.
It is worth noting that the test set is kept unchanged across different category shift scenarios.
We use the per-class accuracy metric as ACC for the generalization performance and measure the AUC score for OOD detection (maximum softmax probability as the confidence). 
Moreover, we denote the harmonic mean of ACC and AUC as HAA, \ie, $\text{HAA} = \frac{2\times \text{ACC} \times \text{AUC}}{\text{ACC} + \text{AUC}}$. 
HAA provides a high score only if the approach performs well in both generalization and OOD detection.

\textbf{Baseline methods.}
We conduct a comparative analysis of UEO against several unsupervised fine-tuning methods, including UPL \citep{huang2022unsupervised} and POUF \citep{tanwisuth2023pouf}, as well as the robust zero-shot inference baseline, CLIP \citep{radford2021learning}.
Besides, we present the results of one modified domain adaptation method, DANCE \citep{saenko2010adapting}, and two model adaptation methods (\ie, Entropy Minimization (EntMin) \citep{wang2021tent} and Mutual Information Maximization (InfoMax) \citep{liang2020we}). 
In DANCE, the source loss is omitted, and we set the trade-off before two target losses as 0.1.

\textbf{Implementation details.}
For all experiments, we utilize the pre-trained ResNet-50 and ViT-B/16 models provided by the official CLIP repository \citep{radford2021learning}.
The epoch number is set to 50 for small-size datasets (\ie, Office and OfficeHome) and 5 for large-size datasets (\ie, VisDA-C and DomainNet), and the learned model in the last epoch is chosen for a fair evaluation.
During training, we use an SGD optimizer with an initial learning rate of 1e-4 for both encoders, except for EntMin, which uses a learning rate of 1e-5.
We also employ a cosine scheduler to gradually decrease the learning rate.
The parameters optimized in all methods include the prompt of the text encoder and affine parameters in the normalization layers (\ie, BatchNorm in ResNet-50 and LayerNorm in ViT-B/16) of the visual encoder.
The context length of the prompt is fixed at 4 and takes the default initialization ``a photo of a".
We reproduce all methods' loss functions using the hyperparameters provided in their respective papers with the experimental results averaged over two different seeds.

\setlength{\tabcolsep}{3.0pt}
\begin{table*}[!ht]
\centering
\caption{Results (\%) on different domains under the open-set category shift (ResNet-50).}
\label{tab:oda}
    \resizebox{0.96\textwidth}{!}{
    \begin{tabular}{l|l|cc>{\columncolor{Gray}}c|cc>{\columncolor{Gray}}c|cc>{\columncolor{Gray}}c|cc>{\columncolor{Gray}}c|cc>{\columncolor{Gray}}c}
    \toprule
        \multirow{2}{*}{\rotatebox[origin=c]{270}{OPT}} & Tasks & \multicolumn{3}{c|}{Office} & \multicolumn{3}{c|}{OfficeHome} & \multicolumn{3}{c|}{VisDA-C} & \multicolumn{3}{c|}{DomainNet} & \multicolumn{3}{c}{Average} \\ 
        & Methods & ACC & AUC & HAA & ACC & AUC & HAA & ACC & AUC & HAA & ACC & AUC & HAA & ACC & AUC & HAA \\ \midrule
        $\times$ & CLIP \citep{radford2021learning} & 73.3 & 82.8 & 77.7 & 73.1 & 75.0 & 73.9 & 88.9 & 81.6 & 85.1 & 47.3 & 66.5 & 53.0 & 64.9 & 74.0 & 67.8 \\ \midrule
        \multirow{6}{*}{\rotatebox[origin=c]{270}{Textual Prompt}}
        & EntMin \citep{wang2021tent} & 73.3 & 82.8 & 77.8 & 73.2 & 75.1 & 74.0 & 88.9 & 81.6 & 85.1 & 46.3 & 65.8 & 51.2 & 64.6 & 73.8 & 67.1 \\ 
        & InfoMax \citep{liang2020we} & 78.3 & 80.3 & 79.2 & 75.6 & 74.6 & 75.0 & 91.2 & 73.8 & 81.4 & 50.5 & 66.3 & 55.3 & 68.2 & 72.3 & 68.8 \\ 
        & DANCE \citep{saito2020universal} & 74.8 & 82.5 & 78.4 & 73.6 & 75.5 & 74.4 & 89.5 & 75.6 & 81.9 & 40.3 & 61.4 & 44.3 & 62.6 & 71.2 & 64.1 \\
        & UPL \citep{huang2022unsupervised} & 76.5 & 81.9 & 79.0 & 75.8 & 75.9 & \textbf{\color{myblue}75.8} & 89.0 & 81.8 & \textbf{\color{myblue}85.3} & 50.3 & 67.4 & 55.8 & 67.5 & 74.5 & 69.7 \\ 
        & POUF \citep{tanwisuth2023pouf} & 76.5 & 82.6 & 79.4 & 74.9 & 75.5 & 75.1 & 91.0 & 78.4 & 84.1 & 49.9 & 66.1 & 54.6 & 67.3 & 73.5 & 69.0 \\ 
        & UEO & 78.1 & 82.7 & \textbf{\color{myblue}80.3} & 76.0 & 74.7 & 75.3 & 91.3 & 79.2 & 84.8 & 50.8 & 67.6 & \textbf{\color{myblue}56.0} & 68.4 & 74.1 & \textbf{\color{myblue}69.8} \\ 
        \midrule
        \multirow{6}{*}{\rotatebox[origin=c]{270}{+ NormLayers}}
        & EntMin \citep{wang2021tent} & 73.7 & 82.8 & 77.9 & 73.8 & 75.4 & 74.5 & 89.8 & 81.6 & 85.5 & 46.5 & 66.0 & 51.4 & 65.0 & 73.9 & 67.4 \\
        & InfoMax \citep{liang2020we} & 78.8 & 80.9 & 79.8 & 77.0 & 74.4 & 75.6 & 92.5 & 77.5 & 84.3 & 51.9 & 65.8 & 56.5 & 69.4 & 72.7 & 70.0 \\
        & DANCE \citep{saito2020universal} & 74.5 & 82.7 & 78.4 & 74.5 & 75.6 & 75.0 & 61.4 & 55.2 & 57.9 & 47.2 & 65.1 & 51.4 & 61.8 & 70.1 & 63.9  \\
        & UPL \citep{huang2022unsupervised} & 76.7 & 82.3 & 79.3 & 76.0 & 76.7 & \textbf{\color{myred}76.3} & 89.1 & 81.8 & 85.3 & 51.2 & 67.4 & 56.7 & 68.0 & 74.8 & 70.2 \\
        & POUF \citep{tanwisuth2023pouf} & 77.0 & 82.8 & 79.7 & 75.8 & 75.4 & 75.6 & 91.8 & 78.7 & 84.6 & 50.1 & 65.1 & 54.9 & 67.9 & 73.2 & 69.4 \\
        & UEO & 77.9 & 82.9 & \textbf{\color{myred}80.3} & 76.8 & 75.0 & 75.9 & 92.6 & 81.2 & \textbf{\color{myred}86.5} & 51.9 & 67.4 & \textbf{\color{myred}57.2} & 69.2 & 74.4 & \textbf{\color{myred}70.7} \\
    \bottomrule
    \end{tabular}
    }
\end{table*}

\setlength{\tabcolsep}{3.0pt}
\begin{table*}[!ht]
\centering
\caption{Results (\%) on different domains under the open-partial-set category shift (ResNet-50).}
\label{tab:opda}
    \resizebox{0.96\textwidth}{!}{
    \begin{tabular}{l|l|cc>{\columncolor{Gray}}c|cc>{\columncolor{Gray}}c|cc>{\columncolor{Gray}}c|cc>{\columncolor{Gray}}c|cc>{\columncolor{Gray}}c}
    \toprule
        \multirow{2}{*}{\rotatebox[origin=c]{270}{OPT}} & Tasks & \multicolumn{3}{c|}{Office} & \multicolumn{3}{c|}{OfficeHome} & \multicolumn{3}{c|}{VisDA-C} & \multicolumn{3}{c|}{DomainNet} & \multicolumn{3}{c}{Average} \\ 
        & Methods & ACC & AUC & HAA & ACC & AUC & HAA & ACC & AUC & HAA & ACC & AUC & HAA & ACC & AUC & HAA \\ \midrule
        $\times$ & CLIP \citep{radford2021learning} & 73.2 & 82.8 & 77.7 & 73.2 & 75.0 & 73.9 & 88.9 & 81.6 & 85.1 & 47.3 & 66.5 & 53.0 & 64.9 & 74.0 & 67.8 \\ \midrule
        \multirow{6}{*}{\rotatebox[origin=c]{270}{Textual Prompt}}
        & EntMin \citep{wang2021tent} & 73.2 & 82.7 & 77.7 & 73.1 & 75.1 & 73.9 & 88.9 & 81.4 & 85.0 & 46.3 & 65.8 & 51.2 & 64.5 & 73.7 & 67.1 \\ 
        & InfoMax \citep{liang2020we} & 76.3 & 81.6 & 78.8 & 75.4 & 74.5 & 74.8 & 84.4 & 67.6 & 74.9 & 50.6 & 66.4 & 55.4 & 66.9 & 71.8 & 67.9 \\ 
        & DANCE \citep{saito2020universal} & 74.2 & 82.9 & 78.3 & 74.5 & 75.2 & 74.8 & 85.8 & 74.1 & 79.4 & 47.9 & 65.2 & 51.9 & 65.3 & 72.6 & 67.0 \\
        & UPL \citep{huang2022unsupervised} & 74.7 & 82.5 & 78.4 & 74.7 & 75.1 & 74.8 & 89.3 & 82.0 & \textbf{\color{myblue}85.5} & 50.0 & 67.3 & 55.6 & 66.8 & 74.4 & \textbf{\color{myblue}69.3} \\ 
        & POUF \citep{tanwisuth2023pouf} & 75.3 & 82.5 & 78.7 & 74.9 & 76.0 & 75.4 & 90.6 & 72.2 & 80.0 & 49.8 & 66.1 & 54.6 & 67.0 & 72.8 & 68.4 \\ 
        & UEO & 76.4 & 83.2 & \textbf{\color{myblue}79.6} & 75.6 & 75.4 & \textbf{\color{myblue}75.5} & 87.5 & 74.7 & 80.5 & 50.7 & 67.6 & \textbf{\color{myblue}55.9} & 67.4 & 73.7 & 69.1 \\ 
        \midrule
        \multirow{6}{*}{\rotatebox[origin=c]{270}{+ NormLayers}}
        & EntMin \citep{wang2021tent} & 73.4 & 82.7 & 77.8 & 73.7 & 75.3 & 74.4 & 89.1 & 80.5 & 88.1 & 46.6 & 66.0 & 51.4 & 64.9 & 73.7 & 67.7 \\
        & InfoMax \citep{liang2020we} & 77.2 & 81.5 & 79.2 & 76.5 & 73.9 & 75.1 & 90.6 & 74.6 & 84.8 & 51.9 & 66.0 & 56.6 & 68.7 & 72.3 & 69.8 \\
        & DANCE \citep{saito2020universal} & 73.5 & 82.5 & 77.8 & 74.6 & 75.2 & 74.8 & 88.5 & 74.8 & \textbf{\color{myred}88.5} & 48.2 & 65.6 & 52.3 & 65.7 & 72.8 & 68.2 \\
        & UPL \citep{huang2022unsupervised} & 75.3 & 82.5 & 78.7 & 74.9 & 75.8 & 75.3 & 89.8 & 82.5 & 88.1 & 51.0 & 67.5 & 56.6 & 67.4 & 74.7 & 70.2 \\
        & POUF \citep{tanwisuth2023pouf} & 75.4 & 82.4 & 78.7 & 75.6 & 75.8 & 75.7 & 91.6 & 74.5 & 87.1 & 50.5 & 65.6 & 55.3 & 67.7 & 72.9 & 69.7 \\
        & UEO & 77.2 & 83.2 & \textbf{\color{myred}80.1} & 76.5 & 75.6 & \textbf{\color{myred}76.0} & 92.0 & 81.2 & 88.0 & 51.9 & 67.4 & \textbf{\color{myred}57.1} & 68.9 & 74.6 & \textbf{\color{myred}70.9} \\
    \bottomrule
    \end{tabular}
    }
\end{table*}

\subsection{Experimental Results}
In Table~\ref{tab:cda}$\sim$\ref{tab:opda}, we evaluate UEO for four distinct category shift scenarios (\ie, closed-set, partial-set, open-set, and open-partial-set) of U$^2$-FT, and also reproduce the baseline methods for fair comparisons.
The first column ``OPT" denotes the optimized parameter space, where ``+ NormLayers" represents optimizing not only the textual prompt but also the affine parameters in normalization layers. 

\textbf{Results under closed-set category shift (Table~\ref{tab:cda}).}
By solely optimizing the textual prompt, UEO achieves the best average results and outperforms all other methods in terms of HAA on 2 out of 4 datasets.
All the methods benefit from the incorporation of optimizing the normalization layers, while UEO still obtains the best HAA value, increasing from 70.2\% to 71.1\%.
In particular, UEO obtains the second-best average scores in ACC and AUC, indicating a good balance between generalization and OOD detection.

\textbf{Results under partial-set category shift (Table~\ref{tab:pda}).}
Under both optimization scenarios, UEO achieves the best average results and consistently ranks within the top two among all the methods in terms of HAA.
Upon careful examination of the values in both ACC and AUC, it is evident that UEO consistently achieves the best performance across both metrics. 
Furthermore, the benefits of incorporating normalization layers are once again confirmed for all the methods.

\setlength{\tabcolsep}{4.0pt}
\begin{table*}[!ht]
\centering
\caption{Results (\%) on the DomainNet dataset under four different category shifts (ViT-B/16, + NormLayers).}
\label{tab:vit}
    \resizebox{0.94\textwidth}{!}{
    \begin{tabular}{l|cc>{\columncolor{Gray}}c|cc>{\columncolor{Gray}}c|cc>{\columncolor{Gray}}c|cc>{\columncolor{Gray}}c|cc>{\columncolor{Gray}}c}
    \toprule
        Tasks & \multicolumn{3}{c|}{closed-set} & \multicolumn{3}{c|}{partial-set} & \multicolumn{3}{c|}{open-set} & \multicolumn{3}{c|}{open-partial-set} & \multicolumn{3}{c}{Average} \\ 
        Methods & ACC & AUC & HAA & ACC & AUC & HAA & ACC & AUC & HAA & ACC & AUC & HAA & ACC & AUC & HAA \\ \midrule
        CLIP \citep{radford2021learning} & 58.2 & 72.6 & 63.0 & 58.2 & 72.6 & 63.0 & 58.2 & 72.6 & 63.0 & 58.2 & 72.6 & 63.0 & 58.2 & 72.6 & 63.0 \\ \midrule
        EntMin \citep{wang2021tent} & 56.0 & 71.6 & 59.3 & 56.0 & 71.7 & 59.3 & 55.7 & 71.3 & 59.0 & 55.8 & 71.4 & 59.1 & 55.9 & 71.5 & 59.2 \\
        InfoMax \citep{liang2020we} & 62.2 & 71.0 & 65.3 & 62.2 & 70.9 & 65.2 & 62.3 & 70.6 & 65.1 & 62.1 & 70.5 & 64.9 & 62.2 & 70.7 & 65.1 \\
        DANCE \citep{saito2020universal} & 57.9 & 72.0 & 60.6 & 57.9 & 72.0 & 60.6 & 57.9 & 71.3 & 60.3 & 57.8 & 71.6 & 60.4 & 57.9 & 71.7 & 60.5 \\
        UPL \citep{huang2022unsupervised} & 61.5 & 72.8 & 65.6 & 61.2 & 73.0 & 65.5 & 61.7 & 72.9 & 65.8 & 61.4 & 73.0 & 65.6 & 61.4 & 72.9 & 65.6 \\
        POUF \citep{tanwisuth2023pouf} & 60.9 & 71.1 & 64.4 & 61.4 & 71.6 & 64.9 & 61.0 & 70.8 & 64.3 & 61.1 & 70.8 & 64.4 & 61.1 & 71.1 & 64.5 \\
        UEO & 62.0 & 72.5 & \textbf{\color{myred}65.7} & 61.9 & 72.5 & \textbf{\color{myred}65.6} & 62.1 & 73.0 & \textbf{\color{myred}66.0} & 62.0 & 72.9 & \textbf{\color{myred}65.8} & 62.0 & 72.7 & \textbf{\color{myred}65.8} \\    
    \bottomrule
    \end{tabular}
    }
\end{table*}

\textbf{Results under open-set category shift (Table~\ref{tab:oda}).}
For the open-set category shift where the training data contains OOD samples, UEO achieves the best average results under both optimization scenarios.
Although UPL also performs well in rejecting OOD samples, it falls short in generalizing to known classes. 
For instance, on the DomainNet dataset and the optimization of both text prompt and visual normalization layers, UEO achieves a better ACC score of 51.9\% with the same AUC score of 67.4\%.

\begin{figure*}[!htb]
\centering
\small
\setlength\tabcolsep{1mm}
\renewcommand\arraystretch{0.1}
\begin{tabular}{cccc}
\includegraphics[width=0.24\linewidth]{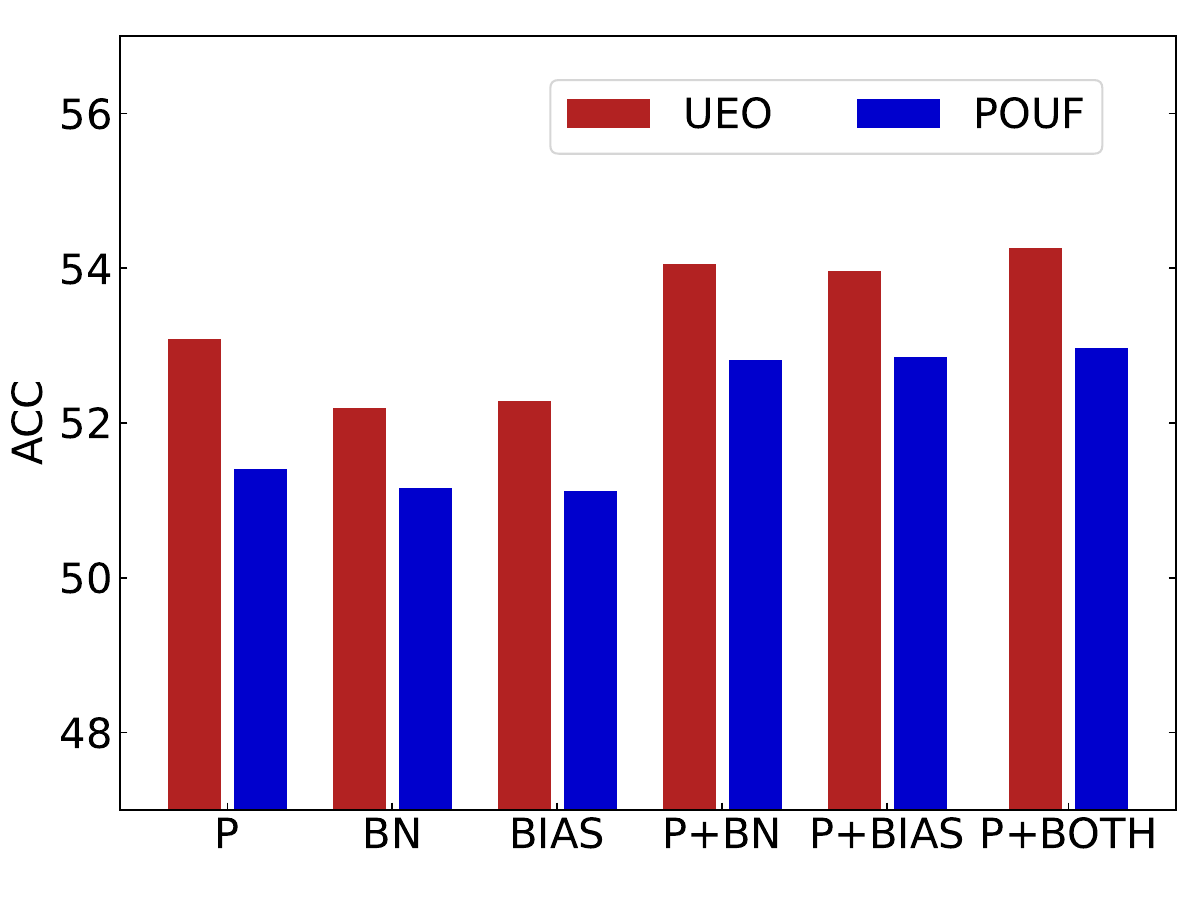} &
\includegraphics[width=0.24\linewidth, clip]{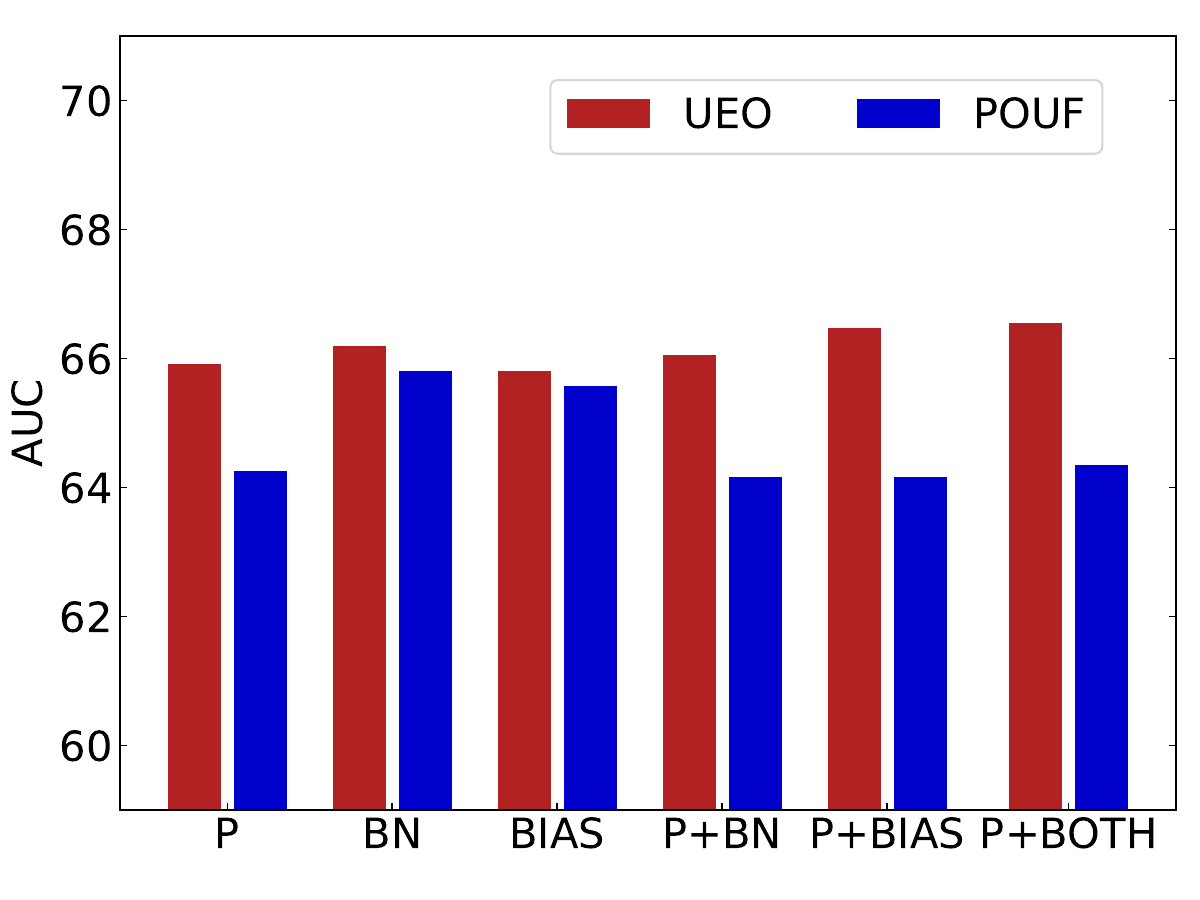} & 
\includegraphics[width=0.24\linewidth, clip]{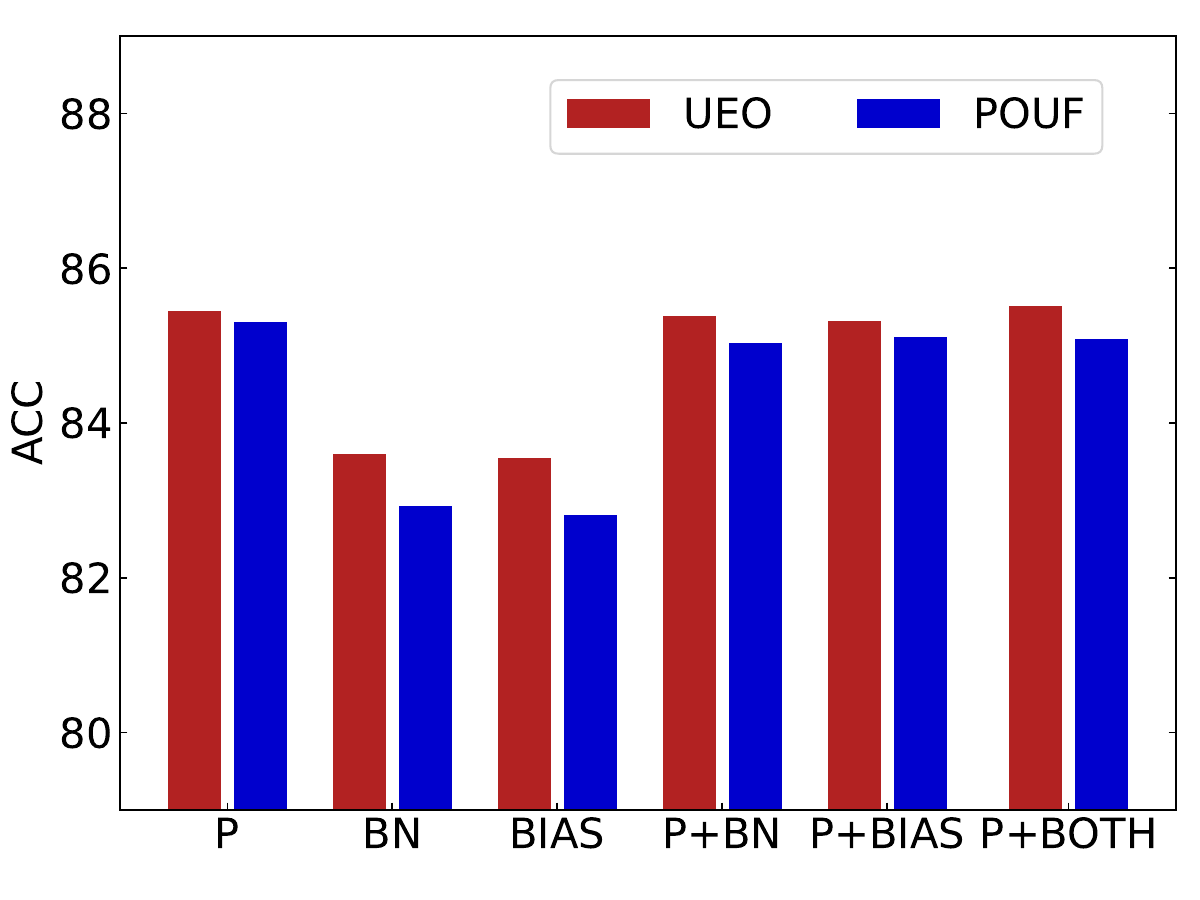} &
\includegraphics[width=0.24\linewidth, clip]{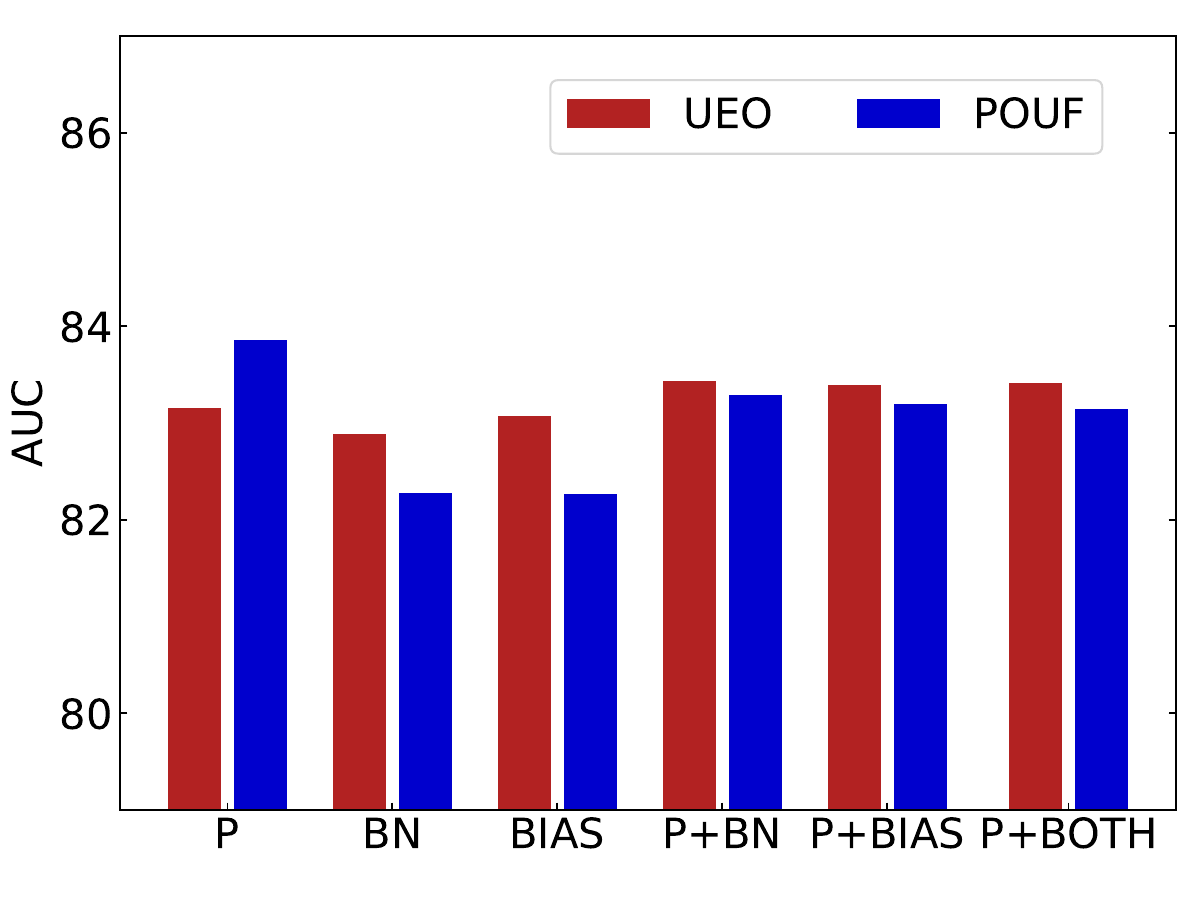} \\
\\[0.5mm]
(a) ACC (\%) on (Sk) & (b) AUC (\%) on (Sk) & (c) ACC (\%) on (Re) & (d) AUC (\%) on (Re)
\end{tabular}
\caption{Different optimization designs on (Sk) in DomainNet and (Re) in OfficeHome, (open-partial-set, ResNet-50).}
\label{fig:opti}
\end{figure*} 

\begin{figure*}[!htb]
\centering
\small
\setlength\tabcolsep{1mm}
\renewcommand\arraystretch{0.1}
\begin{tabular}{cccc}
\includegraphics[width=0.24\linewidth]{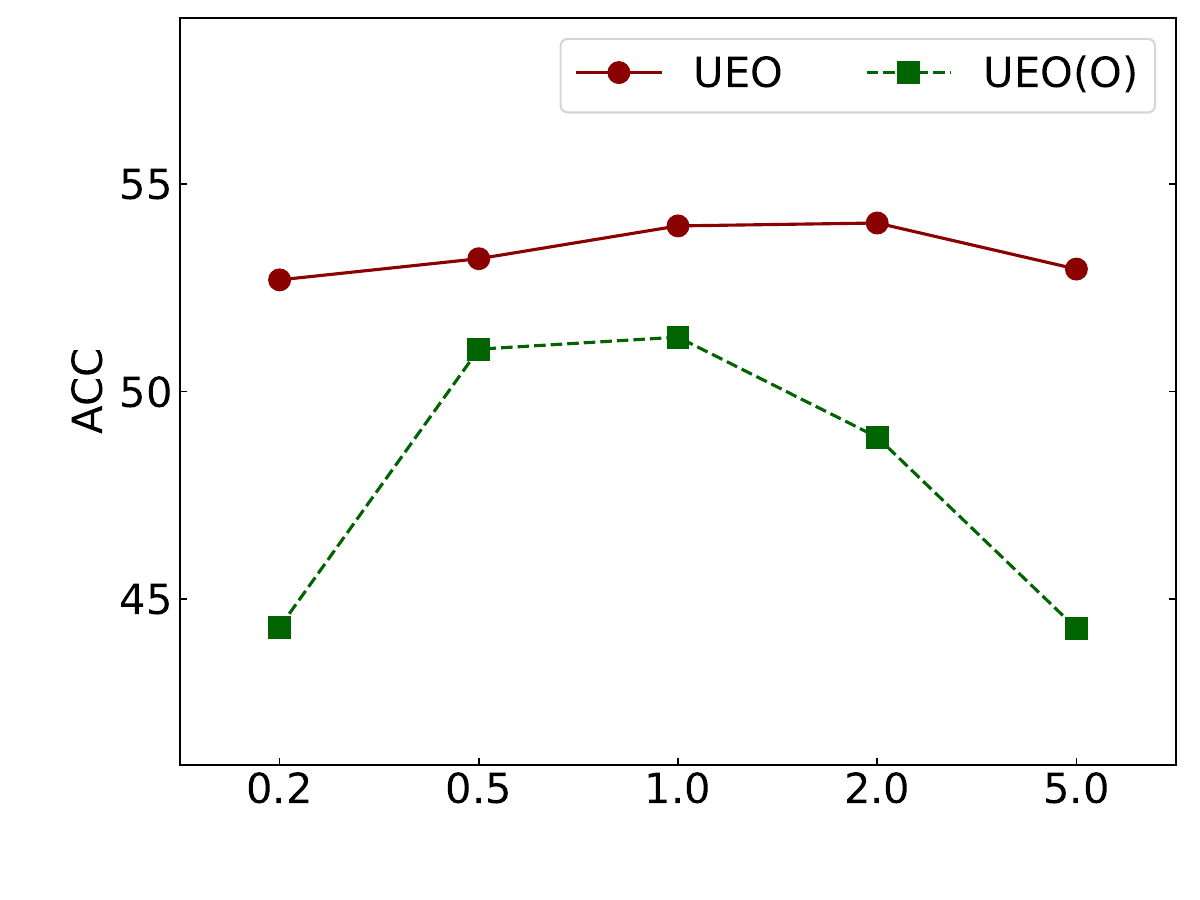} &
\includegraphics[width=0.24\linewidth, clip]{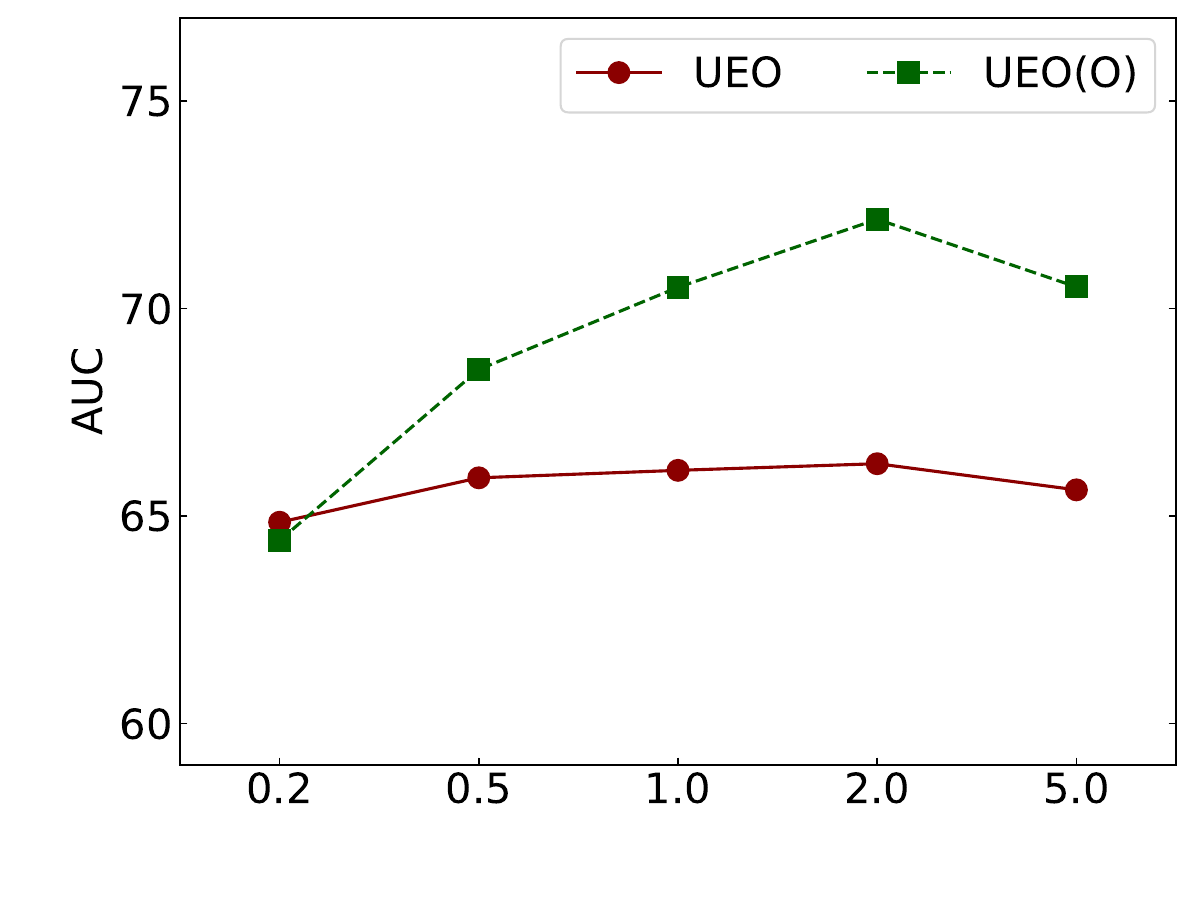} & 
\includegraphics[width=0.24\linewidth, clip]{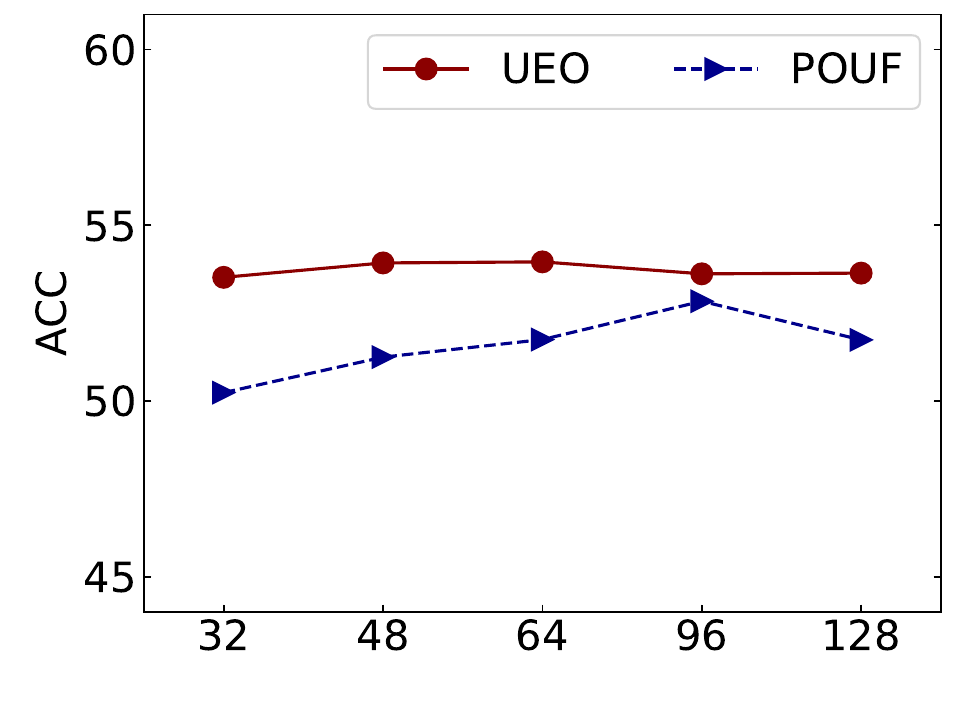} &
\includegraphics[width=0.24\linewidth, clip]{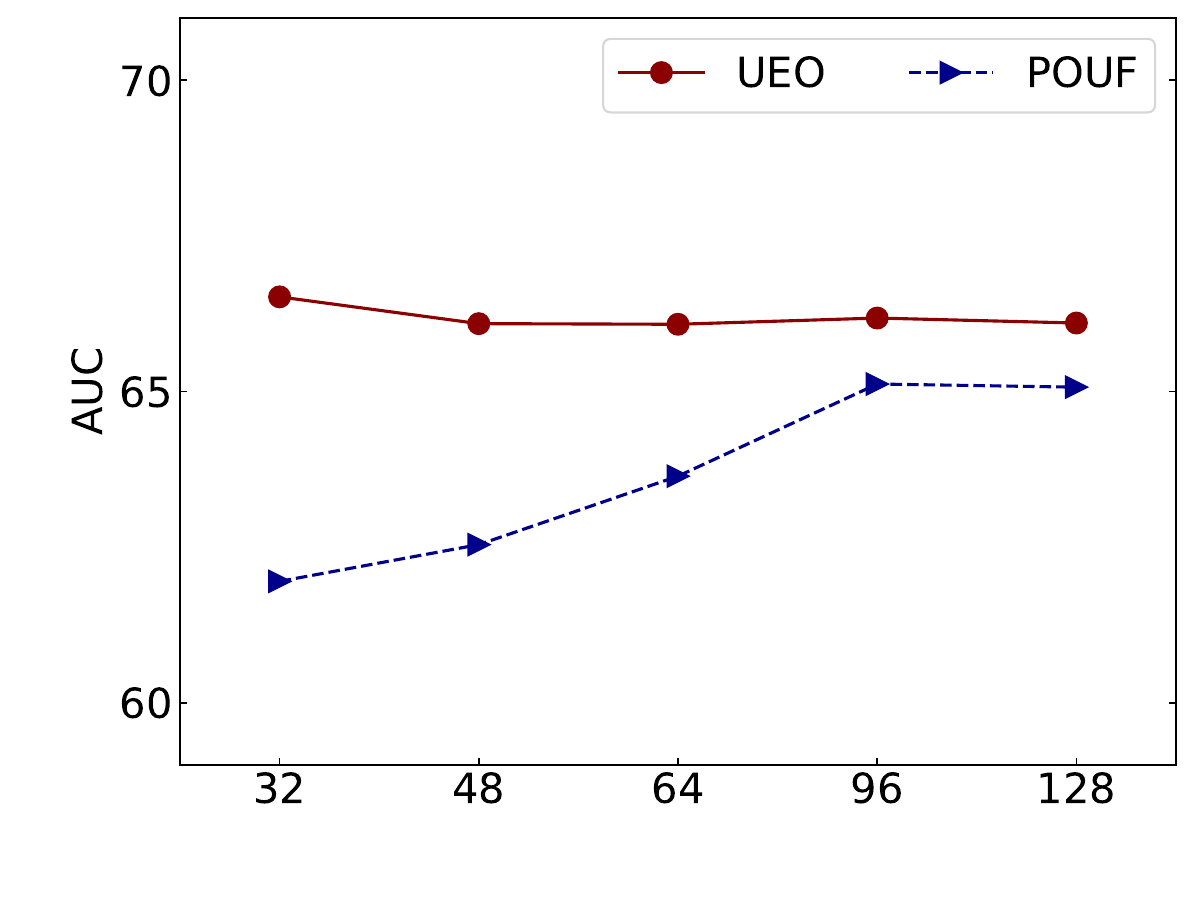} \\
\\[0.5mm]
(a) ACC (\%) \textit{w.r.t.} trade-offs & (b) AUC (\%) \textit{w.r.t.} trade-offs  & (c) ACC (\%) \textit{w.r.t.} batch size  & (d) AUC (\%) \textit{w.r.t.} batch size
\end{tabular}
\caption{Results of different hyperparameters on (Sk) in DomainNet (open-partial-set, ResNet-50).}
\label{fig:sensitivity}
\vspace{-10pt}
\end{figure*} 

\textbf{Results under open-partial-set category shift (Table~\ref{tab:opda}).}
While UEO slightly outperforms UPL when optimizing the textual prompt alone, it surpasses UPL when incorporating normalization layers, achieving the highest HAA value under the open-partial-set category shift.
Typically, UEO consistently strikes a good balance between generalization and OOD detection, enhancing generalization performance without compromising its ability to detect OOD samples.

\subsection{Analysis}
\textbf{ViT backbone.}
We further adopt the ViT-B/16 backbone \citep{radford2021learning} as the pre-trained model and report the results in Table~\ref{tab:vit}.
We observe that UEO achieves the best results of HAA on the DomainNet dataset under all four settings. 
In particular, UEO demonstrates superior OOD detection ability compared to InfoMax, while also achieving better generalization performance compared to UPL.

\textbf{Optimization designs.}
We investigate the performance of UEO and the baseline method (POUF \citep{tanwisuth2023pouf}) under different optimization strategies.
To validate the stability of the training process, we selected 6 combinations of textual prompt (P), BatchNorm (BN), and bias (BIAS) as the test range, according to \citep{zaken2022bitfit}. 
The results depicted in Fig.~\ref{fig:opti} demonstrate that UEO is stable and outperforms POUF in various cases. 
The optimization of the vision encoder leads to better performance, indicating that UEO can benefit from a larger parameter space.

\textbf{Hyperparameter sensitivity.}
Even UEO does not require hyperparameter tuning, we aim to investigate the impact of the second term in Eq.~(\ref{eq:went2}) and introduce a trade-off parameter prior to this term.
We present the results of UEO under varying trade-offs in the loss function and batch size (32, 48, 64, 96, 128) during training in Fig.~\ref{fig:sensitivity}. 
Besides, we introduce an enhanced variant of UEO called UEO(O), where we have knowledge of which training data, i.e., the sample-level weight $w(x)$ equals 1 for ID data and 0 for OOD data.
The results demonstrate that UEO remains stable across changes in the trade-off. 
Although UEO(O) can achieve better detection ability with a larger trade-off, its classification accuracy drops significantly. Additionally, UEO is not sensitive to changes in batch size. 
In contrast, POUF requires a larger batch size to estimate the distribution and performs poorly on smaller batches. 
Therefore, UEO does not suffer from trivial hyperparameter selection.

\begin{table*}[!ht]
\caption{Different choices of $\Phi(\cdot)$ and prompt initialization on (Sk) in DomainNet (open-partial-set, ResNet-50).}
\label{tab:weight_init}
\vspace{5pt}
\renewcommand{\arraystretch}{1.2}
\centering
\resizebox{0.9\linewidth}{!}{
\subfloat[Weight function]{
\begin{tabular}{lcc}
\toprule[0.8pt]
$\Phi(w)$ & ACC & AUC \\
\midrule
$1/w$ & 53.9 & 66.1 \\
$\sqrt{1/w}$ & 54.1 & 66.4 \\
$(1/w)^2$ & 53.7 & 65.9 \\
$1-w$ & 53.9 & 66.4 \\
$\sqrt{1-w}$ & 54.2 & 66.3 \\
$(1-w)^2$ & 53.9 & 66.2 \\
\bottomrule[0.8pt]
\end{tabular}
}
\quad \quad 
\subfloat[Prompt initialization]{
\begin{tabular}{l|cc|cc}
\toprule[0.8pt]
Prompt initialization & ACC (CLIP) & AUC (CLIP) & ACC (UEO) & AUC (UEO) \\
\midrule
`a photo of a' & 49.1 & 65.1 & 54.0 \textbf{\color{myred}(+4.9)} & 66.1 \textbf{\color{myred}(+1.0)} \\
`a photo of many' & 47.9 & 65.0 & 52.9 \textbf{\color{myred}(+5.0)} & 66.3 \textbf{\color{myred}(+1.3)} \\
`a sketch of a' & 52.2 & 65.4 & 54.5 \textbf{\color{myred}(+2.3)} & 65.9 \textbf{\color{myred}(+0.5)} \\
`a painting of a' & 50.3 & 65.5 & 54.0 \textbf{\color{myred}(+3.7)} & 65.7 \textbf{\color{myred}(+0.2)} \\
`this is a photo of a' & 49.3 & 64.9 & 53.2 \textbf{\color{myred}(+3.9)} & 66.1 \textbf{\color{myred}(+1.2)} \\
`this is a picture of a' & 49.2 & 67.4 & 54.3 \textbf{\color{myred}(+5.1)} & 65.9 \textbf{\color{myblue}(-1.5)} \\
\bottomrule[0.8pt]
\end{tabular}
}}
\end{table*}

\textbf{Different choices of $\Phi(\cdot)$ and prompt initialization.}
Table~\ref{tab:weight_init} contain the results of different monotonically decreasing function designs of $\Phi(\cdot)$ in Eq.~(\ref{eq:went2}) and textual prompt initialization under the open-partial-set category shift.
We find that UEO achieves nearly consistent performance under various function designs.
As for prompt initialization, the different designs obtain similar results with the choice (``a photo of a"), achieving better performance for both generalization and OOD detection.

\section{Conclusion}
In this paper, we introduce a novel unsupervised universal fine-tuning (U$^2$-FT) setting for CLIP, which does not rely on prior knowledge about the unlabeled data in the downstream domain.
In addition to assessing the generalization ability to identify samples from the known classes, U$^2$-FT also considers improving the detection of OOD samples after fine-tuning.
We propose Universal Entropy Optimization (UEO), which employs sample-level confidence to approximately minimize the entropy of ID samples and maximize the entropy of OOD samples.
UEO is a simple and parameter-efficient approach that updates only a small number of parameters and does not require any hyper-parameters in the objective. 
Extensive results show that UEO always outperforms existing methods across various category shift scenarios.
We believe that the introduced U$^2$-FT setting is an interesting and important contribution to the field of transfer learning with VLMs and has the potential to attract significant attention.

\textbf{Limitation.}
While we have successfully validated the effectiveness of UEO for image classification, we have not yet studied its application to dense prediction tasks such as segmentation and detection.

\section*{Acknowledgements}
This work was funded by the Beijing Nova Program under Grant Z211100002121108, the National Natural Science Foundation of China under Grant 62276256, and the Young Elite Scientists Sponsorship Program by CAST (2023QNRC001).

\section*{Impact Statement}
This paper contributes to the advancement of the subfield of machine learning, particularly focusing on fine-tuning with CLIP.
While our work holds numerous potential societal implications, we choose not to highlight specific ones here.
As for sensitive applications (e.g., medical diagnosis) or with biased datasets, the proposed method may decrease the OOD detection compared with the original CLIP model.

\bibliography{icml2024}
\bibliographystyle{icml2024}

\newpage
\appendix
\onecolumn

\section{Pseudo code}
\label{app:code}
To facilitate a better understanding of our problem setup and proposed method (UEO), we provide the pseudocode below.
Throughout this paper, we utilize two different scores, namely \textbf{ACC} and \textbf{AUC}, calculated in Line 15 and Line 17, respectively, to evaluate all methods under the U$^2$-FT framework.

\begin{algorithm}
\caption{Universal Entropy Optimization (UEO) for Unsupervised Universal Fine-Tuning (U$^2$-FT)} 
\begin{algorithmic}[1]
\STATE \text{\color{gray}\# The training stage}
\STATE \textbf{Input:} The pre-trained CLIP, unlabeled data $\mathcal{X}_t$ associated with the label set $\textcolor{red}{L_u}$, the name list of known classes $\textcolor{red}{L_p} = \{y_1, \dots, y_C\}$.
\FOR {$epoch=1,2,\ldots$}
\FOR {$iteration=1,2,\ldots$}
\STATE Sample a mini-batch $\mathcal{B}_t$ from $\mathcal{X}_t$.
\STATE Forward $\mathcal{B}_t$ to the \textbf{original} CLIP to obtain the weights $w(x) = \mathcal{S}(x)$ using Eq.~(\ref{eq:msp}).
\STATE Forward $\mathcal{B}_t$ to CLIP to obtain the predictions $\{p(x)\}$ using Eq.~(\ref{eq:clip}).
\STATE Update the parameters of CLIP through the gradient of $\mathcal{L}$ in Eq.~(\ref{eq:went2}).
\ENDFOR
\ENDFOR
\STATE \text{\color{gray}\# The testing stage}
\STATE \textbf{Input:} The adapted CLIP, evaluation data $\mathcal{X}_e$ associated with the label set $\textcolor{red}{L_e}$, the name list of known classes $\textcolor{red}{L_p} = \{y_1, \dots, y_C\}$.
\STATE Split $\mathcal{X}_e$ into two sets, $\mathcal{X}_e^1$ associated with $L_e\cap L_p$ and $\mathcal{X}_e^2$ associated with $L_e\setminus L_p$.
\STATE Forward $\mathcal{X}_e$ to the adapted CLIP to obtain the predictions $\{p(x)\}$ using Eq.~(\ref{eq:clip}).
\STATE Calculate the per-class accuracy (\textbf{ACC}) over $\mathcal{X}_e^1$ based on the argmax operation.
\STATE Obtain the scores $\mathcal{S}(x)$ for both $\mathcal{X}_e^1$ and $\mathcal{X}_e^2$ using Eq.~(\ref{eq:msp}).
\STATE Calculate the \textbf{AUC} score to measure the ability of OOD detection.
\end{algorithmic} 
\end{algorithm}

\section{Analysis beyond open-partial-set category shift}
\begin{figure*}[!ht]
    \centering
    \small
    \setlength\tabcolsep{1mm}
    \renewcommand\arraystretch{0.1}
    \begin{tabular}{cccc}
        \includegraphics[width=0.24\linewidth]{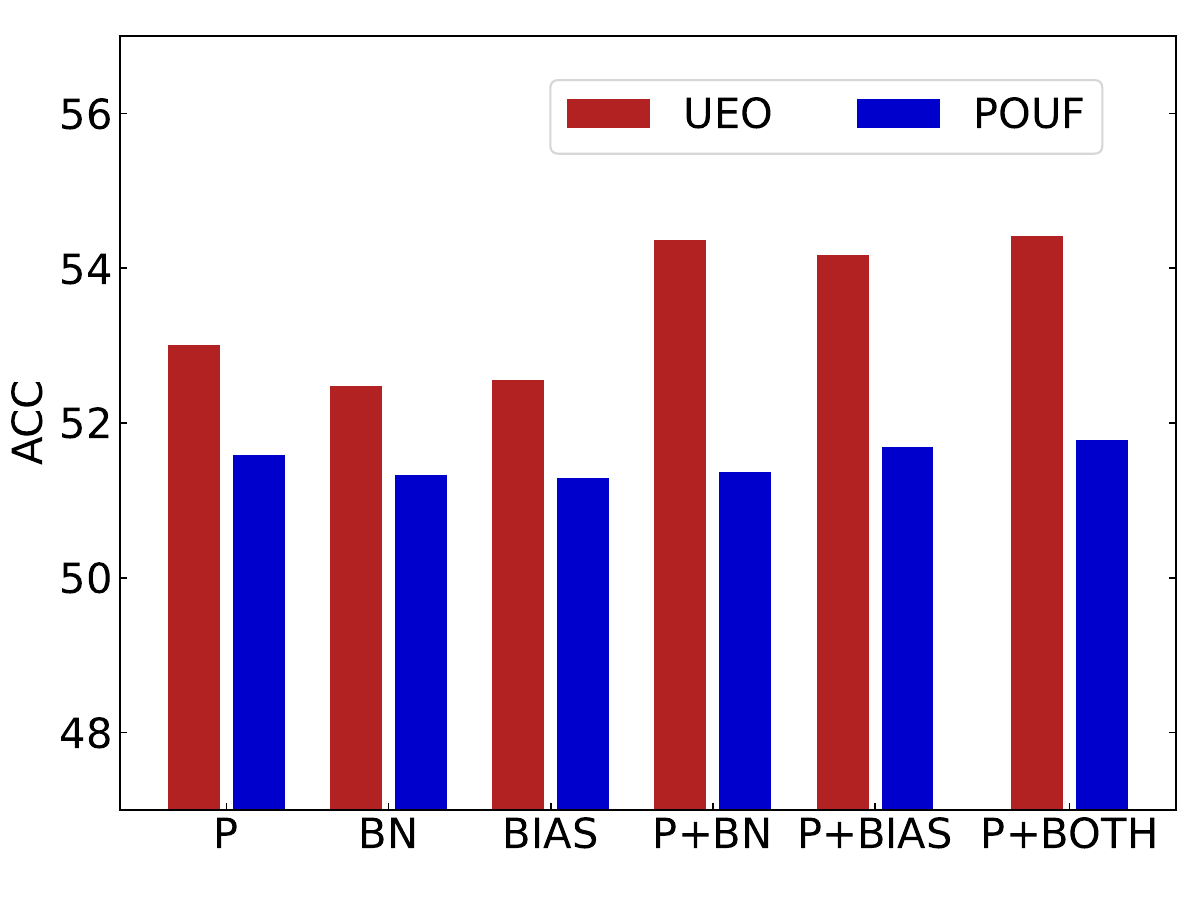} &
        \includegraphics[width=0.24\linewidth, clip]{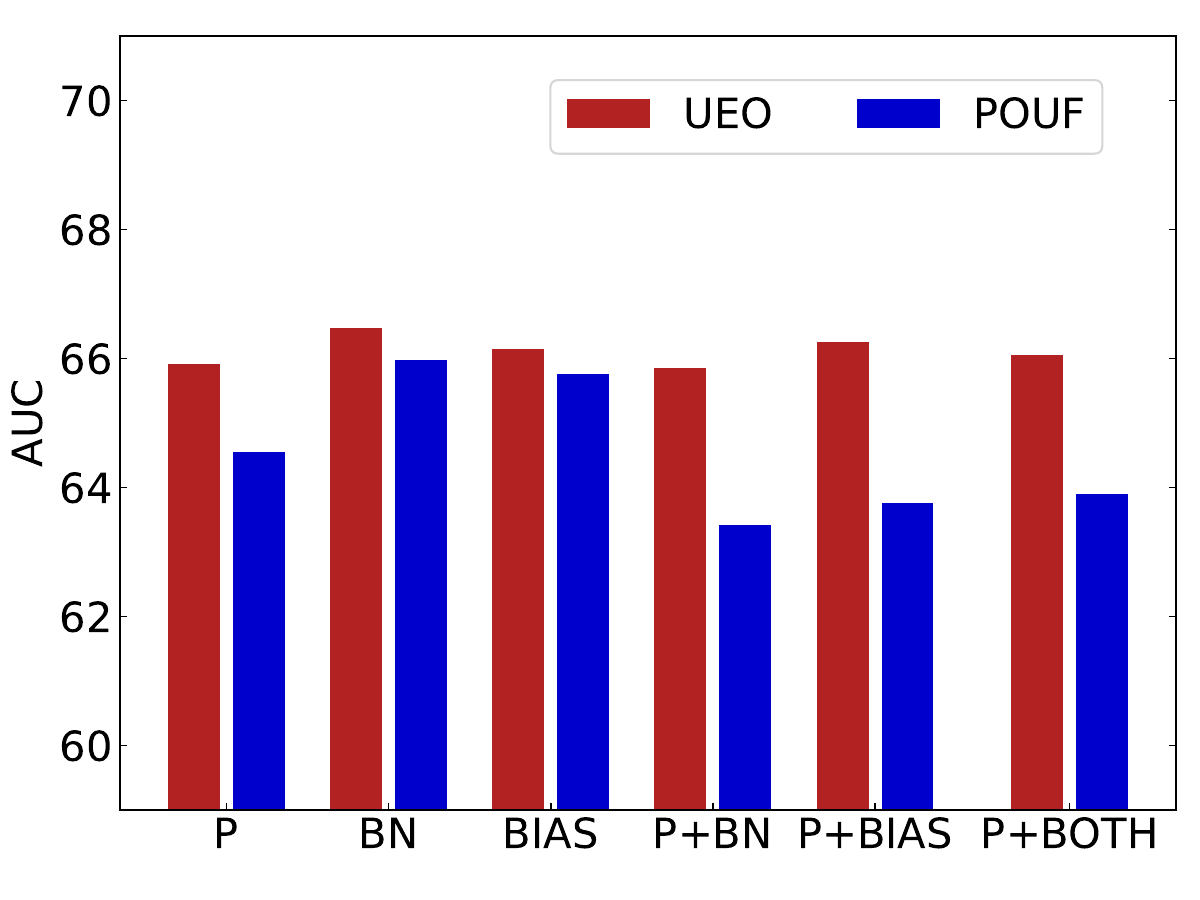} & 
        \includegraphics[width=0.24\linewidth, clip]{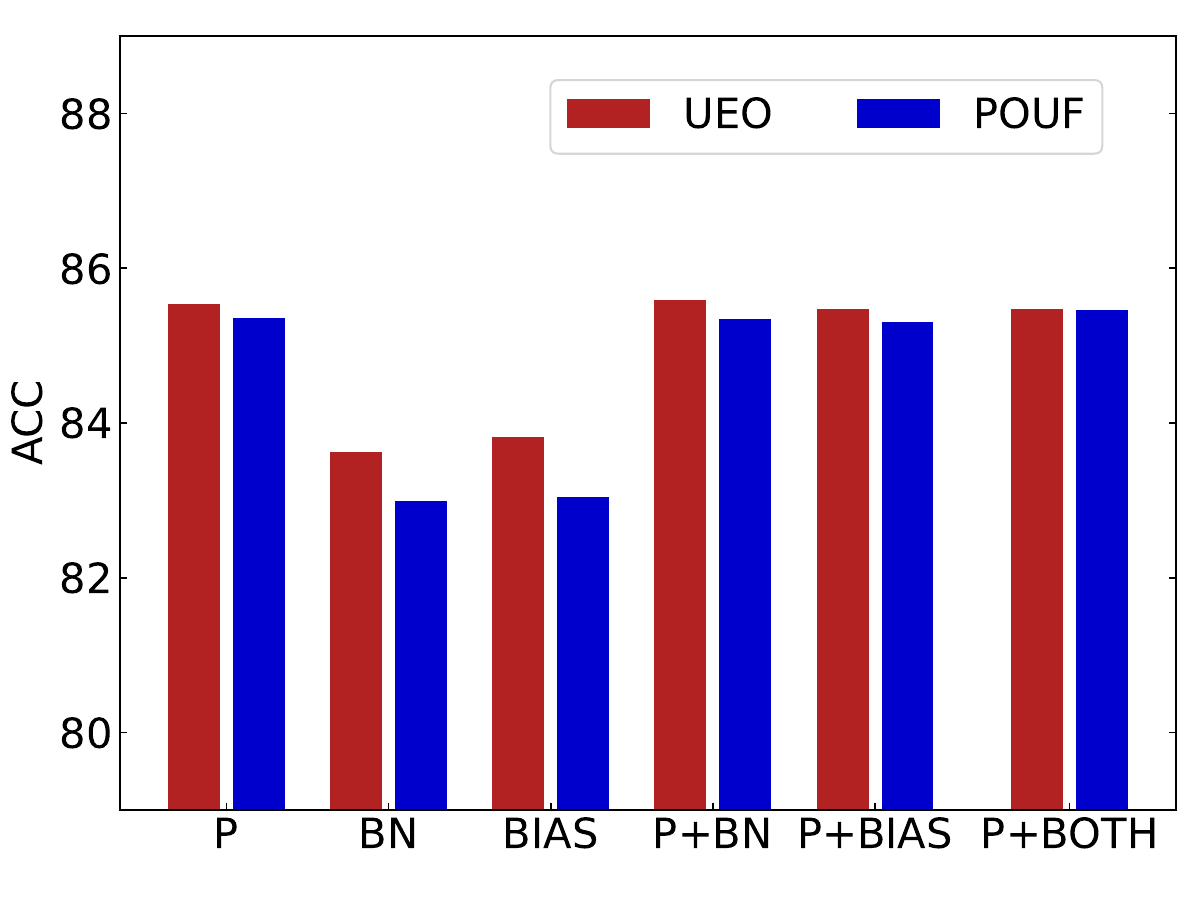} &
        \includegraphics[width=0.24\linewidth, clip]{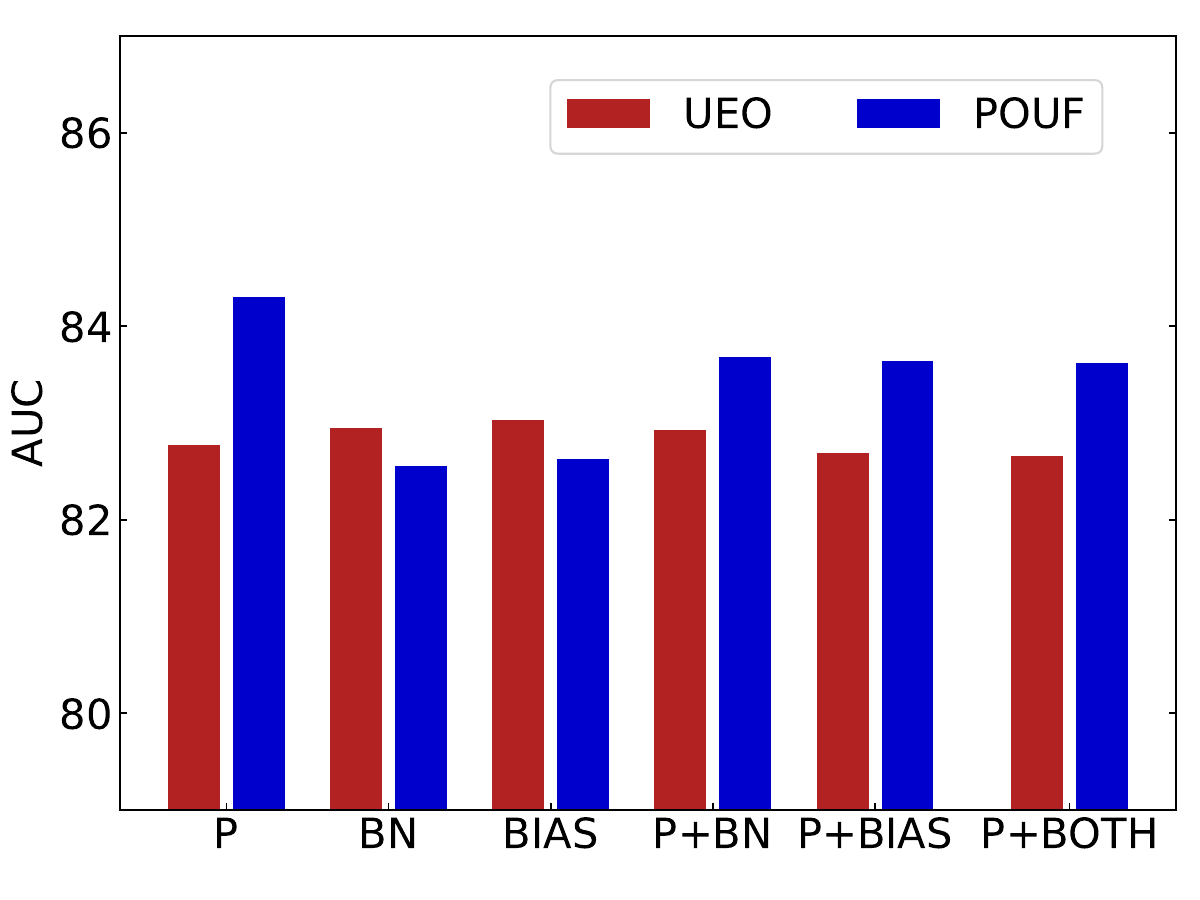} \\
        \\[0.5mm]
        (a) ACC (\%) on (Sk). & (b) AUC (\%) on (Sk). & (c) ACC (\%) on (Re). & (d) AUC (\%) on (Re).
    \end{tabular}
    \caption{Different optimization designs on (Sk) in DomainNet and (Re) in OfficeHome, (closed-set, ResNet-50).}
    \label{fig:opti_cda}
\end{figure*}

\begin{figure*}[!htb]
    \centering
    \small
    \setlength\tabcolsep{1mm}
    \renewcommand\arraystretch{0.1}
    \begin{tabular}{cccc}
        \includegraphics[width=0.24\linewidth]{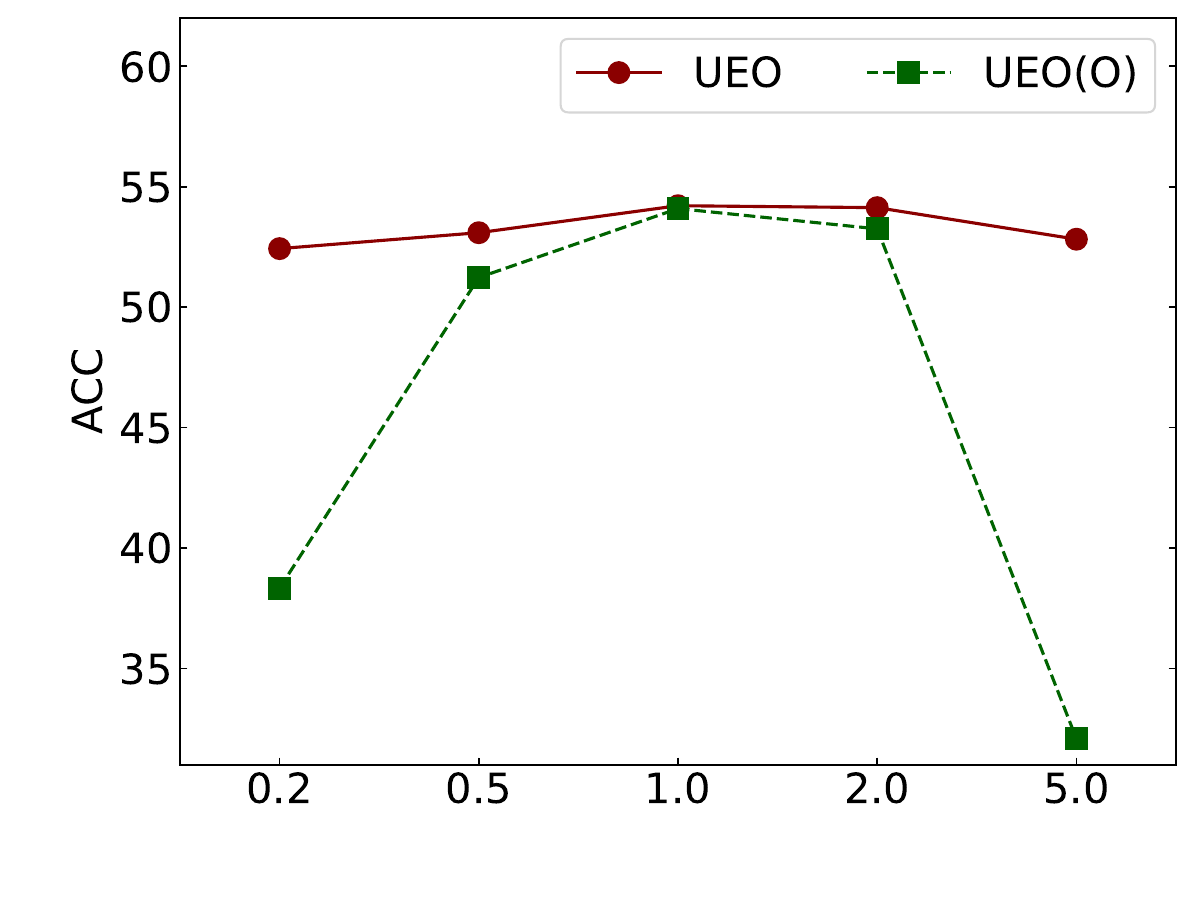} &
        \includegraphics[width=0.24\linewidth, clip]{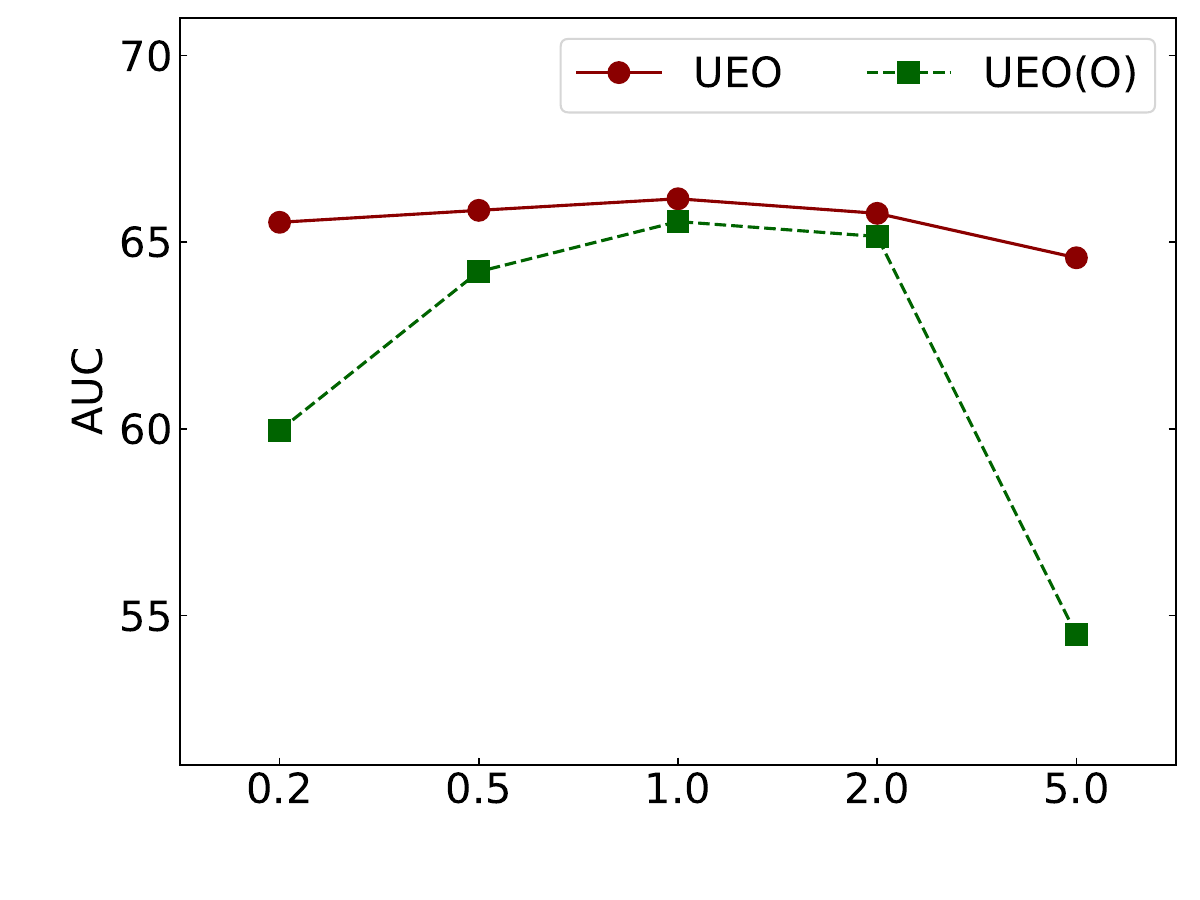} & 
        \includegraphics[width=0.24\linewidth, clip]{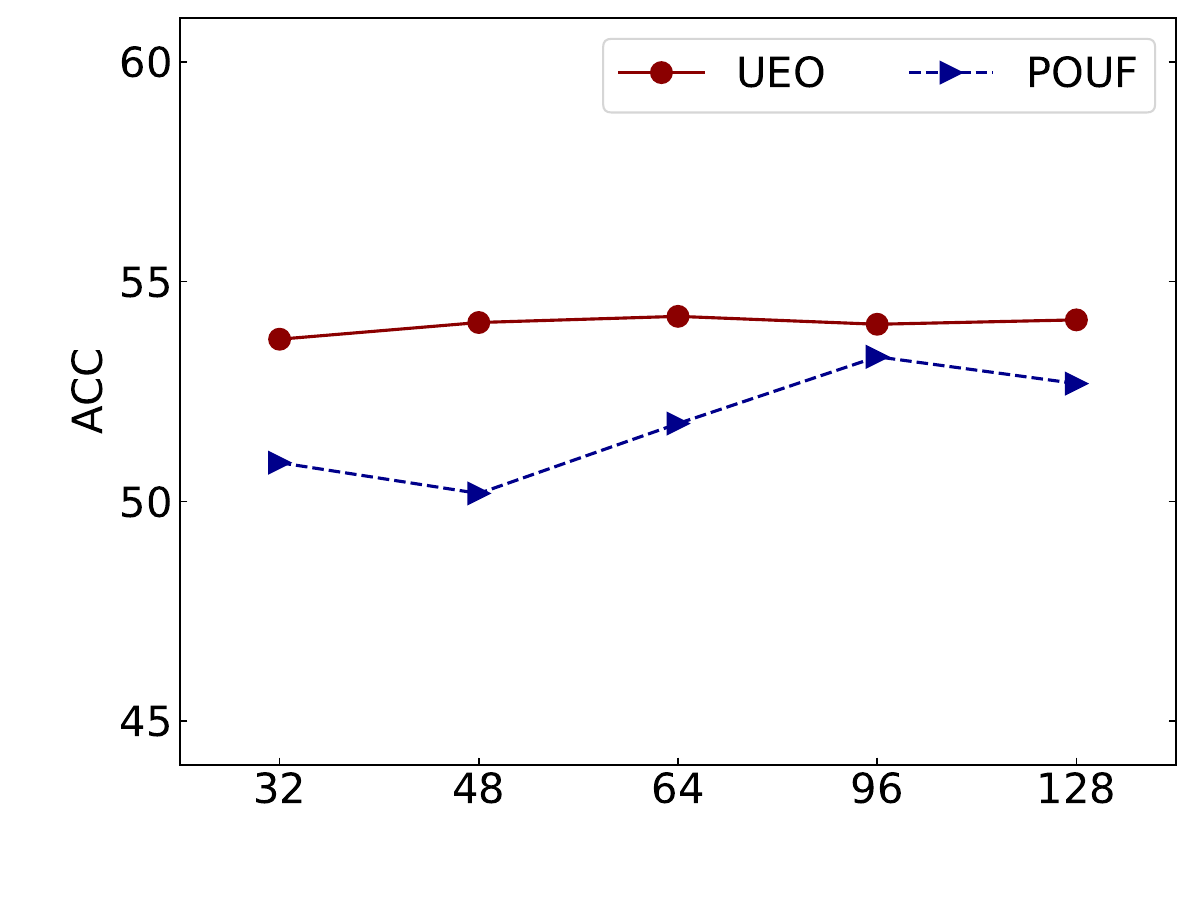} &
        \includegraphics[width=0.24\linewidth, clip]{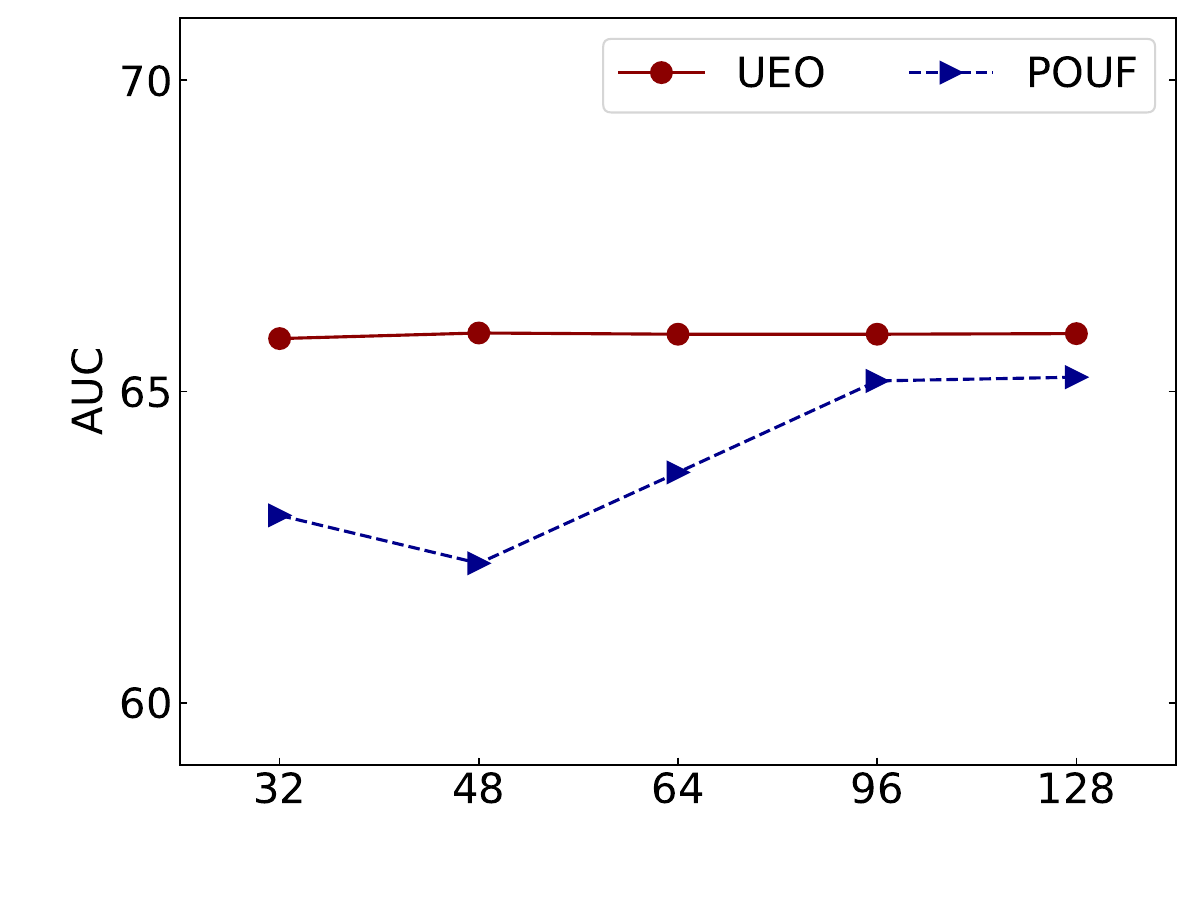} \\
        \\[0.5mm]
        (a) ACC (\%) \textit{w.r.t.} trade-offs & (b) AUC (\%) \textit{w.r.t.} trade-offs & (c) ACC (\%) \textit{w.r.t.} batch size & (d) AUC (\%) \textit{w.r.t.} batch size
    \end{tabular}
    \caption{Results of different hyperparameters on (Sk) in DomainNet (closed-set, ResNet-50).}
    \label{fig:sensitivity_cda}
\end{figure*}

In addition to the primary task analyzed under the open-partial-set category shift in the main text, we also present the results under the closed-set one in Fig.~\ref{fig:opti_cda}, Fig.~\ref{fig:sensitivity_cda}, and Table~\ref{tab:weight_init_cda}, respectively.

\begin{table}[!ht]
\caption{Different choices of $\Phi(\cdot)$ and prompt initialization on (Sk) in DomainNet (closed-set, ResNet-50).}
\label{tab:weight_init_cda}
\vspace{5pt}
\renewcommand{\arraystretch}{1.2}
\centering
\resizebox{0.9\linewidth}{!}{
\subfloat[Weight function]{
\begin{tabular}{lcc}
\toprule[0.8pt]
$\Phi(w)$ & ACC & AUC \\
\midrule
$1/w$ & 54.3 & 65.9 \\
$\sqrt{1/w}$ & 54.2 & 66.0 \\
$(1/w)^2$ & 53.6 & 65.7 \\
$1-w$ & 54.2 & 65.9 \\
$\sqrt{1-w}$ & 54.1 & 66.0 \\
$(1-w)^2$ & 54.1 & 66.0 \\
\bottomrule[0.8pt]
\end{tabular}
}
\quad \quad 
\subfloat[Prompt initialization]{
\begin{tabular}{l|cc|cc}
\toprule[0.8pt]
Prompt initialization & ACC (CLIP) & AUC (CLIP) & ACC(UEO) & AUC (UEO) \\
\midrule
`a photo of a' & 49.1 & 65.1 & 54.4 \textbf{\color{myred}(+5.3)} & 65.9 \textbf{\color{myred}(+0.8)} \\
`a photo of many' & 47.9 & 65.0 & 53.1 \textbf{\color{myred}(+5.2)} & 66.1 \textbf{\color{myred}(+1.1)} \\
`a sketch of a' & 52.2 & 65.4 & 54.6 \textbf{\color{myred}(+2.4)} & 65.8 \textbf{\color{myred}(+0.4)} \\
`a painting of a' & 50.3 & 65.5 & 53.9 \textbf{\color{myred}(+3.6)} & 66.0 \textbf{\color{myred}(+0.5)} \\
`this is a photo of a' & 49.3 & 64.9 & 53.0 \textbf{\color{myred}(+3.7)} & 65.8 \textbf{\color{myred}(+0.9)} \\
`this is a picture of a' & 49.2 & 64.6 & 54.2 \textbf{\color{myred}(+5.0)} & 65.9 \textbf{\color{myred}(+1.3)} \\
\bottomrule[0.8pt]
\end{tabular}
}}
\end{table}

\section{Results under the conventional fine-tuning}
Following the setting of POUF \citep{tanwisuth2023pouf}, we evaluate the effectiveness of UEO under a conventional closed-set setting, which considers all categories as known classes.
From Table~\ref{tab:csda2}, UEO improves the global accuracy on DomainNet from 57.6\% to 61.1\% and outperforms all baseline methods.

\setlength{\tabcolsep}{3.0pt}
\begin{table*}[!ht]
\centering
\caption{Accuracy (\%) under conventional fine-tuning on DomainNet.} 
\label{tab:csda2}
\resizebox{0.85\textwidth}{!}{$
\begin{tabular}{l|cccccc|c|cccccc|c}
\toprule[0.8pt]
Backbones & \multicolumn{7}{c|}{ResNet-50} & \multicolumn{7}{c}{ViT-B/16} \\
\midrule
Methods & Cl & In & Pa & Qu & Re & Sk & Avg. & Cl & In & Pa & Qu & Re & Sk & Avg. \\
\midrule
CLIP \citep{radford2021learning} & 54.8 & 40.9 & 54.6 & 6.0 & 77.7 & 49.2 & 47.2 & 71.0 & 47.6 & 66.2 & 13.9 & 83.7 & 63.5 & 57.6  \\
UPL \citep{huang2022unsupervised} & 58.1 & 45.8 & 56.6 & 11.1 & \textbf{\color{myred}79.6} & 52.8 & 50.7 & 72.4 & 53.9 & 66.8 & \textbf{\color{myred}19.9} & \textbf{\color{myred}84.8} & 65.8 & 60.6 \\
POUF \citep{tanwisuth2023pouf} & 58.9 & 45.8 & 58.1 & 9.6 & 76.6 & 50.7 & 50.0 & 72.3 & 53.3 & \textbf{\color{myred}69.8} & 18.0 & 83.3 & 65.3 & 60.3 \\
UEO & \textbf{\color{myred}59.7} & \textbf{\color{myred}46.6} & \textbf{\color{myred}59.2} & \textbf{\color{myred}11.8} & 78.9 & \textbf{\color{myred}53.8} & \textbf{\color{myred}51.7}  & 73.5 & \textbf{\color{myred}54.2} & 69.2 & 18.9 & 84.5 & \textbf{\color{myred}66.4} & \textbf{\color{myred}61.1} \\
\bottomrule[0.8pt]
\end{tabular}
$}
\end{table*}

\section{Results on general classification datasets}
To validate the effectiveness of the proposed method (UEO), we additionally utilize two widely recognized classification datasets (\ie, ImageNet, and Food101) and present the results under four category shifts in Table~\ref{tab:in}.
The detailed label split is shown in Table~\ref{tab:info}.
In the ImageNet dataset, it is clear to find that UPL and UEO perform better than other methods, and UEO achieves the best score in ACC.
In the Food101 dataset, UEO outperforms nearly all counterparts in both ACC and AUC, achieving the highest HAA value.

\setlength{\tabcolsep}{4.0pt}
\begin{table*}[!ht]
\centering
\caption{Results (\%) on the ImageNet and Food101 datasets under four category shifts (ResNet-50, + NormLayers).}
\label{tab:in}
    \resizebox{0.94\textwidth}{!}{
    \begin{tabular}{l|cc>{\columncolor{Gray}}c|cc>{\columncolor{Gray}}c|cc>{\columncolor{Gray}}c|cc>{\columncolor{Gray}}c|cc>{\columncolor{Gray}}c}
    \toprule
        Tasks & \multicolumn{3}{c|}{closed-set} & \multicolumn{3}{c|}{partial-set} & \multicolumn{3}{c|}{open-set} & \multicolumn{3}{c|}{open-partial-set} & \multicolumn{3}{c}{Average} \\ 
        \textbf{ImageNet} & ACC & AUC & HAA & ACC & AUC & HAA & ACC & AUC & HAA & ACC & AUC & HAA & ACC & AUC & HAA \\ \midrule
        CLIP \citep{radford2021learning} & 63.8 & 73.8 & 68.4 & 63.8 & 73.8 & 68.4 & 63.8 & 73.8 & 68.4 & 63.8 & 73.8 & 68.4 & 63.8 & 73.8 & 68.4 \\ 
        EntMin \citep{wang2021tent} & 64.0 & 73.5 & 68.4 & 64.1 & 73.8 & 68.6 & 63.9 & 73.1 & 68.2 & 63.8 & 73.3 & 68.2 & 63.9 & 73.4 & 68.3 \\
        InfoMax \citep{liang2020we} & 66.3 & 71.8 & 68.9 & 65.4 & 73.4 & 69.2 & 65.1 & 64.5 & 64.8 & 65.3 & 68.3 & 66.8 & 65.5 & 69.5 & 67.4 \\
        DANCE \citep{saito2020universal} & 64.7 & 72.8 & 68.5 & 64.3 & 73.3 & 68.5 & 64.9 & 72.1 & 68.3 & 64.4 & 72.6 & 68.3 & 64.6 & 72.7 & 68.4 \\
        UPL \citep{huang2022unsupervised} & 65.6 & 73.6 & \textbf{\color{myred}69.4} & 65.2 & 73.9 & 69.2 & 65.8 & 73.5 & \textbf{\color{myred}69.4} & 65.6 & 72.9 & 69.1 & 65.5 & 73.5 & \textbf{\color{myred}69.3} \\
        POUF \citep{tanwisuth2023pouf} & 65.7 & 72.0 & 68.7 & 65.2 & 72.8 & 68.8 & 65.7 & 71.0 & 68.3 & 65.8 & 71.7 & 68.6 & 65.6 & 71.9 & 68.6 \\
        UEO & 65.9 & 73.0 & 69.3 & 65.3 & 74.2 & \textbf{\color{myred}69.4} & 65.9 & 72.4 & 69.0 & 65.9 & 73.2 & \textbf{\color{myred}69.3} & 65.7 & 73.2 & 69.2 \\
        \midrule
        Tasks & \multicolumn{3}{c|}{closed-set} & \multicolumn{3}{c|}{partial-set} & \multicolumn{3}{c|}{open-set} & \multicolumn{3}{c|}{open-partial-set} & \multicolumn{3}{c}{Average} \\ 
        \textbf{Food101} & ACC & AUC & HAA & ACC & AUC & HAA & ACC & AUC & HAA & ACC & AUC & HAA & ACC & AUC & HAA \\ 
        \midrule
        CLIP \citep{radford2021learning} & 79.7 & 77.6 & 78.6 & 79.7 & 77.6 & 78.6 & 79.7 & 77.6 & 78.6 & 79.7 & 77.6 & 78.6 & 79.7 & 77.6 & 78.6 \\ 
        EntMin \citep{wang2021tent} & 80.1 & 77.6 & 78.8 & 79.9 & 77.6 & 78.7 & 80.0 & 77.2 & 78.5 & 79.9 & 77.1 & 78.5 & 80.0 & 77.4 & 78.6 \\
        InfoMax \citep{liang2020we} & 82.5 & 77.1 & 79.7 & 82.1 & 77.7 & 79.9 & 82.6 & 76.5 & 79.4 & 82.3 & 76.0 & 79.0 & 82.4 & 76.8 & 79.5 \\
        DANCE \citep{saito2020universal} & 81.0 & 77.4 & 79.2 & 80.4 & 77.4 & 78.9 & 80.4 & 76.3 & 78.3 & 80.0 & 76.2 & 78.0 & 80.4 & 76.8 & 78.6 \\
        UPL \citep{huang2022unsupervised} & 80.2 & 77.7 & 78.9 & 80.6 & 77.3 & 78.9 & 80.3 & 78.0 & 79.1 & 81.2 & 78.4 & 79.8 & 80.6 & 77.8 & 79.2 \\
        POUF \citep{tanwisuth2023pouf} & 81.5 & 77.6 & 79.5 & 81.1 & 77.6 & 79.3 & 81.5 & 77.2 & 79.3 & 81.3 & 77.2 & 79.2 & 81.3 & 77.4 & 79.3 \\
        UEO & 82.4 & 77.9 & \textbf{\color{myred}80.1} & 81.9 & 77.9 & \textbf{\color{myred}79.8} & 82.3 & 77.4 & \textbf{\color{myred}79.8} & 82.1 & 77.7 & \textbf{\color{myred}79.9} & 82.2 & 77.7 & \textbf{\color{myred}79.9} \\   
    \bottomrule
    \end{tabular}
    }
\end{table*}

\section{Fine-tuning Beyond CLIP}
We further investigate the effectiveness of UEO using a well-trained visual classifier like that in the source-free domain adaptation topic \citep{liang2020we,huang2021model}.
In particular, we train a classification model following \citep{liang2020we} using the labeled data in the source domain, then provide it to the unlabeled data in the wild.
We employ two transfer tasks in Office and OfficeHome and show the results of UEO and other methods in Table~\ref{tab:vm}.
It could be easily found that UEO outperforms all counterparts in both metrics, achieving a strong result in terms of HAA for each category shift.

\setlength{\tabcolsep}{4.0pt}
\begin{table*}[!ht]
\centering
\caption{Results (\%) using a well-trained visual model under four different category shifts (ResNet-50, + NormLayers).}
\label{tab:vm}
    \resizebox{0.94\textwidth}{!}{
    \begin{tabular}{l|cc>{\columncolor{Gray}}c|cc>{\columncolor{Gray}}c|cc>{\columncolor{Gray}}c|cc>{\columncolor{Gray}}c|cc>{\columncolor{Gray}}c}
    \toprule
        Tasks & \multicolumn{3}{c|}{closed-set} & \multicolumn{3}{c|}{partial-set} & \multicolumn{3}{c|}{open-set} & \multicolumn{3}{c|}{open-partial-set} & \multicolumn{3}{c}{Average} \\ 
        Office (D$\to$A) & ACC & AUC & HAA & ACC & AUC & HAA & ACC & AUC & HAA & ACC & AUC & HAA & ACC & AUC & HAA \\ \midrule
        Fine-tuned ResNet-50 & 65.2 & 71.8 & 68.3 & 65.2 & 71.8 & 68.3 & 65.2 & 71.8 & 68.3 & 65.2 & 71.8 & 68.3 & 65.2 & 71.8 & 68.3 \\ 
        EntMin \citep{wang2021tent} & 64.3 & 72.0 & 68.0 & 64.9 & 72.8 & 68.6 & 63.2 & 69.8 & 66.3 & 63.9 & 70.5 & 67.0 & 64.1 & 71.3 & 67.5 \\
        InfoMax \citep{liang2020we} & 71.8 & 77.1 & 74.4 & 70.1 & 74.5 & 72.2 & 72.2 & 74.2 & 73.2 & 69.4 & 70.8 & 70.1 & 70.9 & 74.2 & 72.5 \\
        DANCE \citep{saito2020universal} & 65.0 & 72.5 & 68.5 & 66.9 & 73.5 & 70.0 & 65.8 & 73.2 & 69.3 & 64.8 & 71.8 & 68.1 & 65.6 & 72.7 & 69.0 \\
        UPL \citep{huang2022unsupervised} & 69.7 & 73.8 & 71.7 & 67.7 & 72.2 & 69.9 & 69.9 & 74.0 & 71.9 & 68.1 & 70.0 & 69.0 & 68.8 & 72.5 & 70.6 \\
        POUF \citep{tanwisuth2023pouf} & 69.9 & 75.7 & 72.7 & 68.9 & 73.7 & 71.2 & 70.1 & 74.2 & 72.1 & 68.5 & 71.7 & 70.1 & 69.4 & 73.8 & 71.5 \\
        UEO & 71.6 & 75.5 & \textbf{\color{myred}73.5} & 71.1 & 74.9 & \textbf{\color{myred}73.0} & 71.6 & 75.0 & \textbf{\color{myred}73.3} & 70.2 & 74.1 & \textbf{\color{myred}72.1} & 71.1 & 74.9 & \textbf{\color{myred}73.0} \\ 
        \midrule
        Tasks & \multicolumn{3}{c|}{closed-set} & \multicolumn{3}{c|}{partial-set} & \multicolumn{3}{c|}{open-set} & \multicolumn{3}{c|}{open-partial-set} & \multicolumn{3}{c}{Average} \\ 
        OfficeHome (Pr$\to$Ar) & ACC & AUC & HAA & ACC & AUC & HAA & ACC & AUC & HAA & ACC & AUC & HAA & ACC & AUC & HAA \\ \midrule
        Fine-tuned ResNet-50 & 54.6 & 64.5 & 59.1 & 54.5 & 64.5 & 59.1 & 54.6 & 64.5 & 59.1 & 54.5 & 64.5 & 59.1 & 54.6 & 64.5 & 59.1 \\ 
        EntMin \citep{wang2021tent} & 52.5 & 64.3 & 57.8 & 53.2 & 64.4 & 58.2 & 51.5 & 62.6 & 56.5 & 52.1 & 62.4 & 56.8 & 52.3 & 63.4 & 57.3 \\
        InfoMax \citep{liang2020we} & 57.9 & 65.4 & 61.4 & 56.9 & 65.5 & 60.9 & 57.2 & 63.8 & 60.3 & 57.1 & 63.6 & 60.2 & 57.3 & 64.5 & 60.7 \\
        DANCE \citep{saito2020universal} & 53.9 & 65.2 & 59.0 & 54.2 & 65.0 & 59.1 & 53.2 & 62.7 & 57.5 & 54.0 & 64.7 & 58.9 & 53.8 & 64.4 & 58.6 \\
        UPL \citep{huang2022unsupervised} & 57.0 & 65.3 & 60.9 & 55.6 & 65.8 & 60.3 & 56.2 & 65.7 & 60.6 & 55.4 & 64.9 & 59.8 & 56.1 & 65.4 & 60.4 \\
        POUF \citep{tanwisuth2023pouf} & 57.1 & 64.6 & 60.6 & 56.1 & 64.6 & 60.0 & 57.2 & 63.6 & 60.2 & 56.4 & 63.4 & 59.7 & 56.7 & 64.0 & 60.2 \\
        UEO & 58.3 & 66.3 & \textbf{\color{myred}62.0} & 58.2 & 66.2 & \textbf{\color{myred}61.9} & 57.9 & 64.9 & \textbf{\color{myred}61.2} & 58.2 & 64.7 & \textbf{\color{myred}61.3} & 58.1 & 65.5 & \textbf{\color{myred}61.6} \\   
    \bottomrule
    \end{tabular}
    }
\end{table*}

\section{Information of data split for different shifts}
\label{app:split}
In this section, we present specific information about $L_p$ (the target class of interest), $L_u$ (the label set of the training data), and $L_e$ (the label set of the evaluation data) for all datasets.

\setlength{\tabcolsep}{5.0pt}
\begin{table}[htb]
\centering
\caption{Detailed information about four category shifts in the training stage and evaluation stage.}
\label{tab:info}
\resizebox{0.7\textwidth}{!}{$
\begin{tabular}{lllll}
\toprule[0.8pt]
Datasets & Category Shifts & $L_p$ & $L_e$ & $L_u$ \\
\midrule
\multirow{4}{*}{Office} & closed-set & $[0, 25)$ & $[0, 31)$ & $[0, 25)$ \\
& partial-set & $[0, 25)$ & $[0, 31)$ & $[0, 15)$ \\
& open-set & $[0, 25)$ & $[0, 31)$ & $[0, 28)$ \\
& open-partial-set & $[0, 25)$ & $[0, 31)$ & $[0, 15) \cup [25, 28)$ \\
\midrule
\multirow{4}{*}{OfficeHome} & closed-set & $[0, 50)$ & $[0, 65)$ & $[0, 50)$ \\
& partial-set & $[0, 50)$ & $[0, 65)$ & $[0, 35)$ \\
& open-set & $[0, 50)$ & $[0, 65)$ & $[0, 60)$ \\
& open-partial-set & $[0, 50)$ & $[0, 65)$ & $[0, 35) \cup [50, 60)$ \\
\midrule
\multirow{4}{*}{VisDA-C} & closed-set & $[0, 8)$ & $[0, 12)$ & $[0, 8)$ \\
& partial-set & $[0, 8)$ & $[0, 12)$ & $[0, 6)$ \\
& open-set & $[0, 8)$ & $[0, 12)$ & $[0, 10)$ \\
& open-partial-set & $[0, 8)$ & $[0, 12)$ & $[0, 6) \cup [8, 10)$ \\
\midrule
\multirow{4}{*}{DomainNet} & closed-set & $[0, 300)$ & $[0, 345)$ & $[0, 300)$ \\
& partial-set & $[0, 300)$ & $[0, 345)$ & $[0, 250)$ \\
& open-set & $[0, 300)$ & $[0, 345)$ & $[0, 330)$ \\
& open-partial-set & $[0, 300)$ & $[0, 345)$ & $[0, 250) \cup [300, 330)$ \\
\midrule
\multirow{4}{*}{ImageNet} & closed-set & $[0, 600)$ & $[0, 1000)$ & $[0, 600)$ \\
& partial-set & $[0, 600)$ & $[0, 1000)$ & $[0, 400)$ \\
& open-set & $[0, 600)$ & $[0, 1000)$ & $[0, 800)$ \\
& open-partial-set & $[0, 600)$ & $[0, 1000)$ & $[0, 400) \cup [600, 800)$ \\
\midrule
\multirow{4}{*}{Food101} & closed-set & $[0, 60)$ & $[0, 101)$ & $[0, 60)$ \\
& partial-set & $[0, 60)$ & $[0, 101)$ & $[0, 40)$ \\
& open-set & $[0, 60)$ & $[0, 101)$ & $[0, 80)$ \\
& open-partial-set & $[0, 60)$ & $[0, 101)$ & $[0, 40) \cup [60, 80)$ \\
\bottomrule[0.8pt]
\end{tabular}
$}
\end{table}

\end{document}